\documentclass[a4paper,twoside]{article}

\usepackage{epsfig}
\usepackage{subcaption}
\usepackage{calc}
\usepackage{amssymb}
\usepackage{amstext}
\usepackage{amsmath}
\usepackage{amsthm}
\usepackage{multicol}
\usepackage{pslatex}
\usepackage{apalike}
\usepackage{algorithm2e}
\usepackage[bottom]{footmisc}
\usepackage{SCITEPRESS}     %

\usepackage[inline]{enumitem}
\usepackage{graphicx}
\usepackage{color}
\usepackage{booktabs}
\usepackage{siunitx}
\usepackage[colorlinks]{hyperref}
\usepackage{cleveref}
\usepackage[rgb]{xcolor}
\usepackage{placeins}

\usepackage{tikz}
\usepackage{pgfplots}

\usepgflibrary{fpu}
\usepgfplotslibrary{fillbetween}

\usetikzlibrary{patterns}
\usetikzlibrary{positioning}
\usetikzlibrary{calc}
\definecolor{meancolor}{HTML}{feb24c}
\definecolor{stdcolor}{HTML}{ffeda0}
\definecolor{fccolor}{HTML}{b2df8a}
\definecolor{outcolor}{HTML}{065b7c}
\definecolor{conccolor}{HTML}{a6cee3}
\definecolor{convcolor}{HTML}{ff5733}

\usepackage{stfloats}

\begin{document}

\title{Can Bayesian Neural Networks Explicitly Model Input Uncertainty?}

\author{\authorname{Matias Valdenegro-Toro\sup{1}\orcidAuthor{0000-0001-5793-9498} and Marco Zullich\sup{1}\orcidAuthor{0000-0002-9920-9095}}
    \affiliation{\sup{1}Department of Artificial Intelligence, University of Groningen, Nijenborgh 9, 9747AG, Groningen, Netherlands}
\email{m.a.valdenegro.toro@rug.nl, m.zullich@rug.nl}
}

\abstract{Inputs to machine learning models can have associated noise or uncertainties, but they are often ignored and not modelled. It is unknown if Bayesian Neural Networks and their approximations are able to consider uncertainty in their inputs. In this paper we build a two input Bayesian Neural Network (mean and standard deviation) and evaluate its capabilities for input uncertainty estimation across different methods like Ensembles, MC-Dropout, and Flipout. Our results indicate that only some uncertainty estimation methods for approximate Bayesian NNs can model input uncertainty, in particular Ensembles and Flipout.}

\onecolumn \maketitle \normalsize \setcounter{footnote}{0} \vfill

\section{\uppercase{Introduction}}
\label{sec:introduction}

In the last two decades, Neural Networks (NNs) have become state-of-the-art applications in many different domains, such as computer vision and natural language processing.
Despite this, these models are known as being notoriously bad at modelling uncertainty, especially when considering the \emph{frequentist} setting \cite{valdenegro2021find}, in which fixed parameters are \emph{trained} to minimize a loss function.
Indeed, while NNs for regression lack a direct way to estimate uncertainty, Deep NNs for classification are often found to be extremely overconfident in their predictions \cite{guo2017calibration}, even when running inference with random data \cite{nguyen2015deep}.
Bayesian Neural Networks (BNNs), which consider parameters as probability distributions, provide a natural way to produce uncertainty estimates, both in the regression and in the classification setting \cite{papadopoulos2001confidence}.
They have been proven to be substantially better at producing more reliable uncertainty estimates, albeit the quality depends on the techniques which are used to approximate these models \cite{ovadia2019trust}.
A model whose uncertainty estimates are reliable is also called \emph{calibrated}, and one of the main metrics for calibration is called the Expected Calibration Error (ECE) \cite{naeini2015obtaining}.

\begin{figure}[t]
    \centering
    \includegraphics[width=0.825\linewidth]{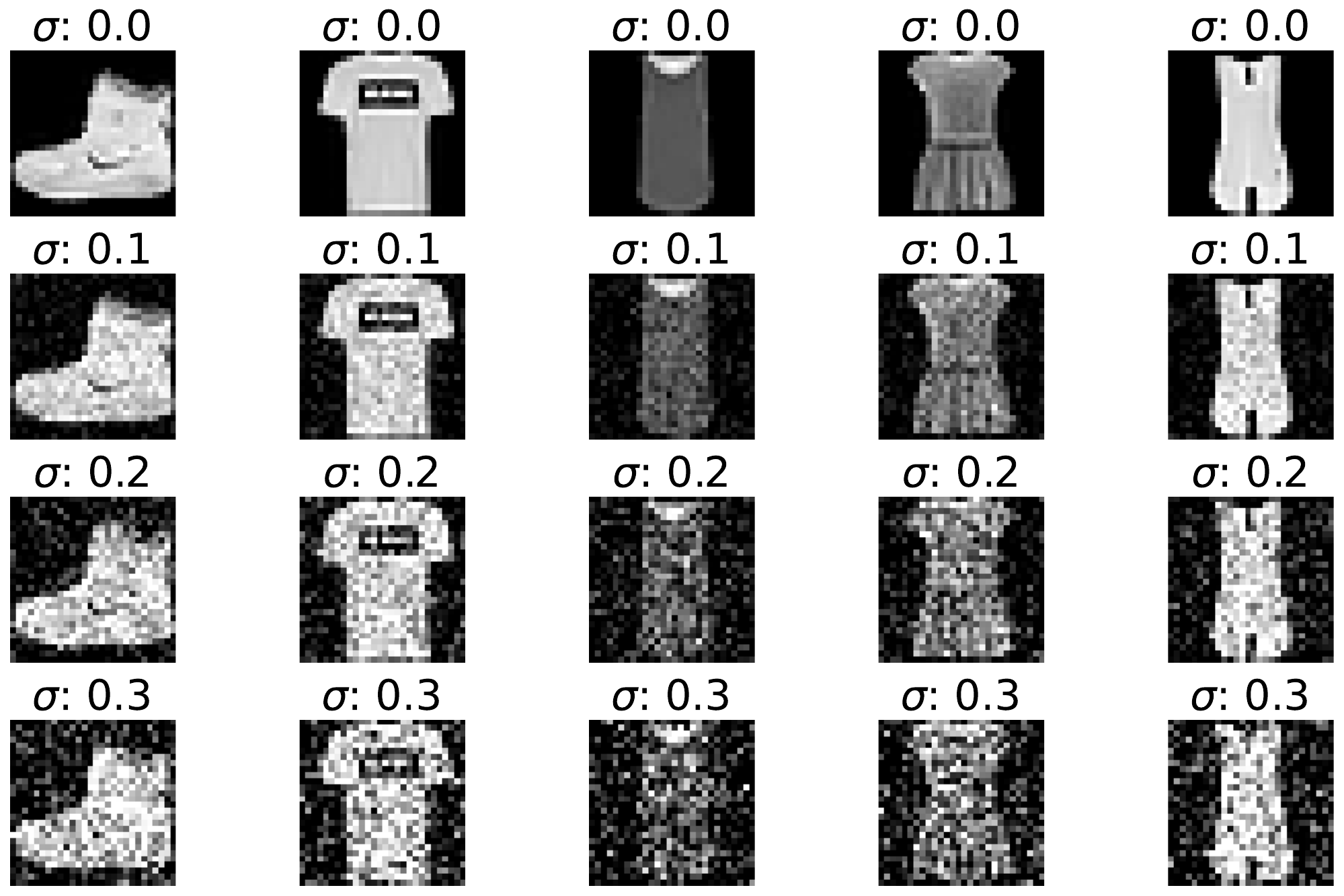}
    \caption{
        Sample of data from the Fashion-MNIST dataset with Gaussian noise with increasing standard deviation ($\sigma$ in the figure) added.
        The first row ($\sigma=0.0$) represents the original, unperturbed data.
        Natural data are often captured by means of digital sensors, which are prone to be noisy and can sporadically fail.
        Training NNs which can effectively model input uncertainty, especially when the noise is anomalously high, is important in having reliable predictions, which can be discarded whenever the predictive uncertainty of the model is too high.
    }
    \vspace*{2em}
    \label{fig:fmnist}
\end{figure}

\begin{figure*}[t]
    \centering
    \includegraphics[width=\linewidth]{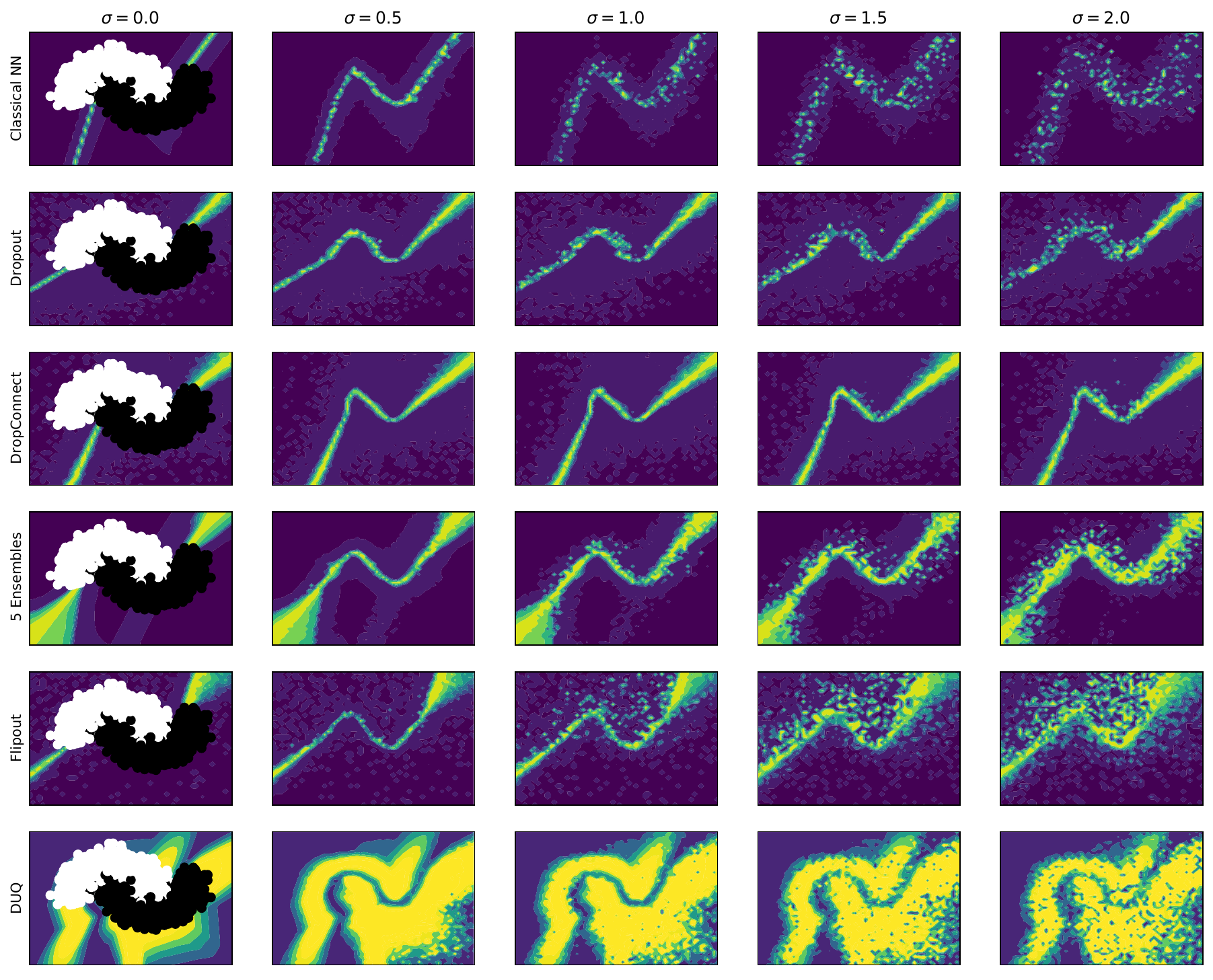}
    \caption{Comparison on the Two Moons dataset with training $\sigma = 0.2$, as the testing standard deviation is varied. Each heatmap indicates predictive entropy (low blue to high yellow) and the first column includes the training data points, With larger test standard deviation, some UQ methods do not significantly change their output uncertainty (DropConnect, Dropout, DUQ), while Flipout and Ensembles do have significant changes, indicating that they are able to model input uncertainty and propagate it to the output.}
    \label{fig:two_moons_comparison}
    \vspace*{1em}
\end{figure*}

Uncertainty, in the context of Machine Learning, is split in two categories \cite{hullermeier2021aleatoric}:
\begin{enumerate*}[label=(\alph*)]
    \item \emph{epistemic} or \emph{model} uncertainty, and
    \item \emph{aleatoric} or \emph{data} uncertainty---here also called \emph{input} uncertainty.
\end{enumerate*}
These two types of uncertainty are usually implicitly modelled together in a single concept, called \emph{predictive} uncertainty, and the process of recovering the epistemic and aleatoric components is called uncertainty \emph{disentanglement} \cite{valdenegro2022deeper}.
An effective modeling of aleatoric and epistemic uncertainty by a machine learning model is crucial whenever this model needs to
\begin{enumerate*}[label=(\alph*)]
    \item be deployed in-the-wild, or
    \item be used (in assisting) for decision-making in safety-critical situations.
\end{enumerate*}
In these cases, it is paramount that model is well calibrated: if it is presented with anomalous or unknown data---for which it effectively behaves randomly---we want this reflected in the prediction uncertainty.
In this sense, a highly unconfident prediction can be discarded \emph{a priori} because it has a high chance of being inaccurate.

The digitization of natural data requires capturing it with either manual measurements or sensors, both procedures which are subjects to noise: this represents aleatoric uncertainty; recapturing the data within the same condition several times will lead to different measurements.
We can thus summarize each data point as a \emph{mean} data and the corresponding \emph{standard deviation}.

In the present work, instead of letting the NNs implicitly model predictive uncertainty, we provide the input uncertainty as \emph{input}, in addition to the \emph{mean} value of the data.
We call these models \emph{two-input NNs}.

We provide our results on two small-scale classification tasks: the Two Moons toy example and the Fashion-MNIST dataset \cite{xiao2017fashion}.
We train classical NNs and five approximate classes of BNNs (MC-Dropout, MC-DropConnect, Ensembles, Flipout, Direct Uncertainty Quantification---DUQ) on these tasks.
We observe the behavior of the uncertainty and ECE when different values of noise are injected into the data and conclude that often these models fail to correctly estimate input uncertainty.

\begin{figure*}[th]
    \centering
    \begin{subfigure}{0.19\linewidth}
        \includegraphics[width=0.9\linewidth]{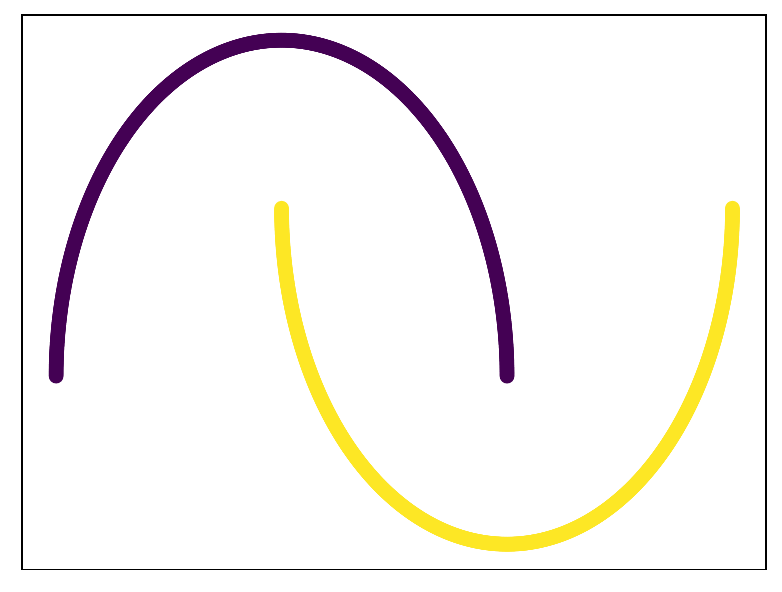}
        \caption{$\sigma = 0.0$}
    \end{subfigure}
    \begin{subfigure}{0.19\linewidth}
        \includegraphics[width=0.9\linewidth]{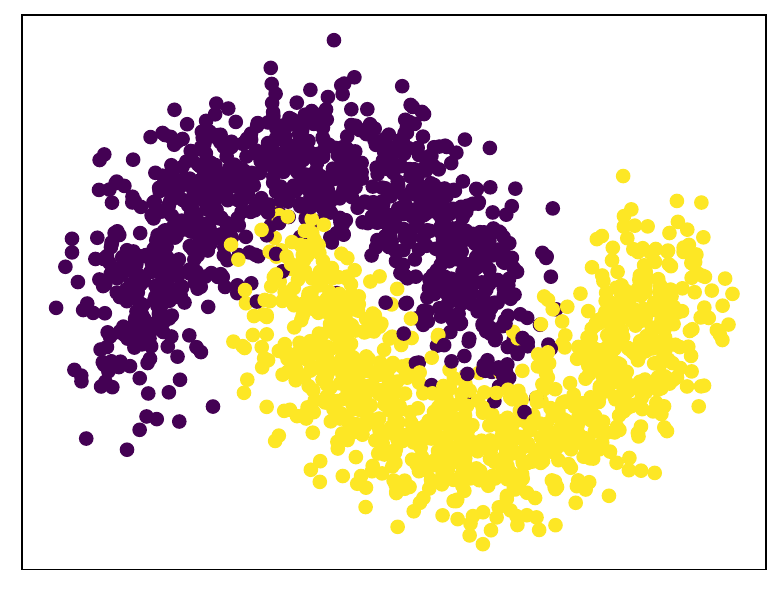}
        \caption{$\sigma = 0.2$}
    \end{subfigure}
    \begin{subfigure}{0.19\linewidth}
        \includegraphics[width=0.9\linewidth]{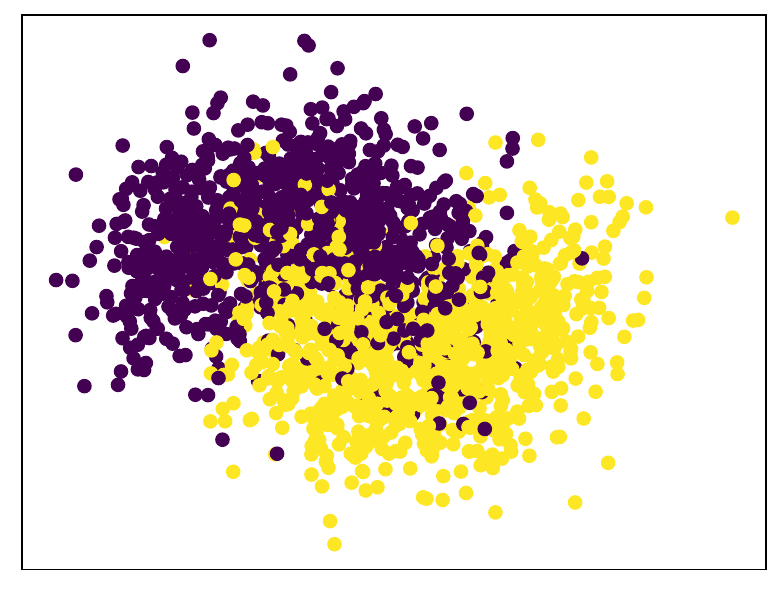}
        \caption{$\sigma = 0.4$}
    \end{subfigure}
    \begin{subfigure}{0.19\linewidth}
        \includegraphics[width=0.9\linewidth]{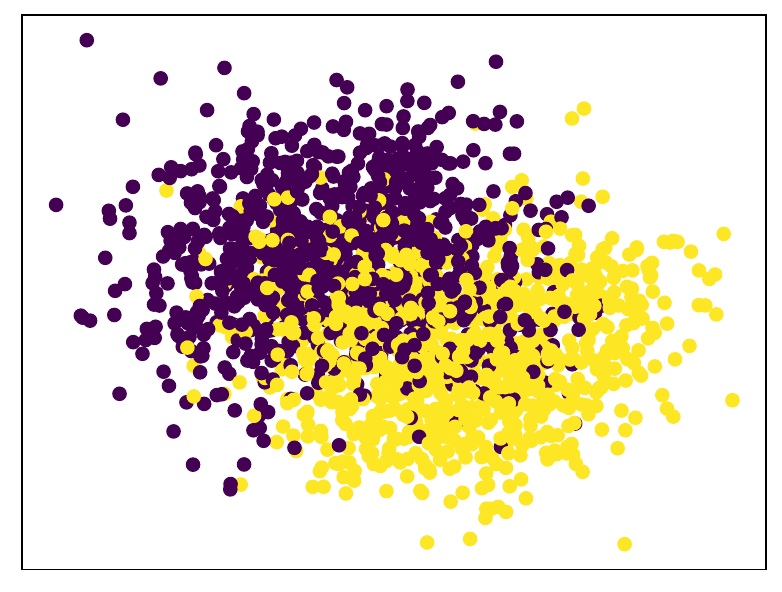}
        \caption{$\sigma = 0.6$}
    \end{subfigure}
    \begin{subfigure}{0.19\linewidth}
        \includegraphics[width=0.9\linewidth]{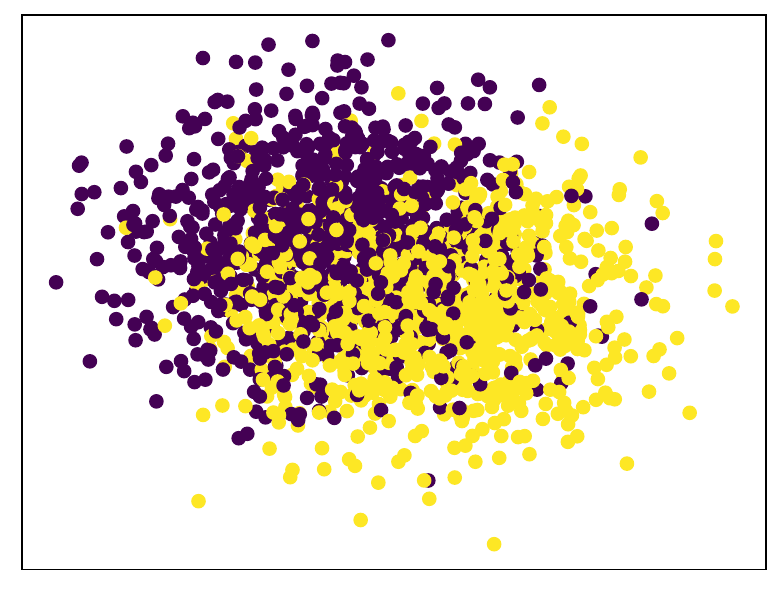}
        \caption{$\sigma = 0.8$}
    \end{subfigure}
    \vspace*{1em}
    \caption{The version of the Two Moons dataset (with \num{1000} data points) used in the present work, the two colors representing the two categories.
        From left to right, we add an increasingly higher level of zero-mean Gaussian noise. 
        The standard deviation is denoted by $\sigma$.}    
    \vspace*{1em}
    \label{fig:2moons_dataset}
\end{figure*}

The investigation of the quality of predictive uncertainty estimates for machine learning models is a long-studied subject and is usually associated with Bayesian modeling \cite{roberts1965probabilistic}: given the fact that these models output a probability distribution, its deviation can be used as a natural estimate of uncertainty.
Deterministic NNs predict point estimates, thus they lack a natural expression of uncertainty, except for the case of classification, where the output---after the application of the softmax function---is interpretable as a probability distribution.
Initial attempts at computation of probability intervals on the output of NNs include the usage of \emph{two-headed models} which output a mean prediction and the standard deviation \cite{nix1994estimating} and early-day BNNs \cite{mackay1992evidence}.
None of these attempts, though, propose a direct modeling of data uncertainty.

Nonetheless, there are more recent efforts for achieving this.
\cite{wright1998neural} and \cite{wright1999bayesian} use the Laplace approximation to train a BNN with input uncertainty, but this is not a modern BNN and it is only tested on simplistic regression settings.
\cite{tzelepis2017linear} introduce a variation of Support Vector Machines which include a Gaussian noise formulation for each data point, which is directly taken into consideration in the hinge loss for determining the separating hyperplane.
\cite{rodrigues2023information} introduce the concept of two-input NNs by crafting a simple toy classification problem, showing how, by providing more information as input to the models, their NNs perform better than the regular, ``single-input'' counterparts.
\cite{hullermeier2014learning}, instead, focuses on producing \emph{fuzzy} loss functions to utilize in a deterministic setting. This allows to incorporate input uncertainty in the empirical risk minimization paradigm.
All of these three works limit their investigations to toy problems and, moreover, do not provide insights into the evaluation of uncertainty estimates.

To the best of our knowledge, we are the first to investigate the capability of modern BNNs to explicitly model input uncertainty, by providing an analysis on the quality of the uncertainty that these models produce.
Our hypothesis is that models being presented with aleatoric uncertainty as input will not be able to effectively reflect it in the predictive uncertainty, exhibiting high levels of confidence even when the input is anomalously noisy.

The contributions of this work are an evaluation of the capability of BNNs to explicitly model uncertainty in their inputs, we evaluate several uncertainty estimation methods and approximate BNNs, and conclude that only Ensembles are---to a certain extent---reliable when considering explicit uncertainty in its input.

\section{Evaluating Bayesian Neural Networks against Input Uncertainty}

\subsection{Datasets}\label{sec:datasets}
We base our experiments on two datasets, the Two Moons dataset and Fashion-MNIST.

\textbf{Two Moons}.
Two Moons is a toy binary classification problem available in the Python library scikit-learn \cite{kramer2016scikit}.
It is composed by a variable number of 2d data points generated in forming two interleaving half circles.
Due to the ease of visualization, it is often being used in research on uncertainty estimation for visualizing the capability of the models to produce reliable uncertainty values in and around the domain of the dataset.
Notice that, due to its toy nature, this dataset only comes with a training set, i.e., there are no validation or test splits.
Some examples of unperturbed and perturbed Two Moons dataset with \num{1000} data points are visible in \Cref{fig:2moons_dataset}.

\textbf{Fashion-MNIST}.
Fashion-MNIST is a popular benchmark for image classification introduced by \cite{xiao2017fashion} as a more challenging version of MNIST \cite{lecun1998gradient}.
It features \num{70000} grayscale, $28\times 28$ images of clothing items from 10 different categories.
The images come pre-split into a training set of \num{60000} and a test set of \num{10000} images.
A sample of unperturbed and perturbed images from Fashion-MNIST is showcased in \Cref{fig:fmnist}. 

\begin{figure*}[t]
    \centering
    \hspace*{-0.5cm} 
    \begin{tikzpicture}[scale=0.8, every node/.style={draw, circle, minimum size=0.4cm}]
        
        \node[draw=none, fill=none, rotate=90] at (-0.5, 2.2) {mean $x_\mu$};
        \node[fill=meancolor] (inputmean1) at (0, 2.6) {};
        \node[fill=meancolor] (inputmean2) at (0, 1.8) {};
        
        \node[draw=none, fill=none, rotate=90] at (-0.5, 0.4) {std $x_\sigma$};
        \node[fill=stdcolor] (inputstd1) at (0, 0.8){};
        \node[fill=stdcolor] (inputstd2) at (0, 0) {};
        
        \draw[draw, rounded corners=5pt, fill=fccolor] (1, 1.7) rectangle ++(2, 1) node[midway, draw=none, fill=none, align=center] (fc1_1) {FC layer\\10 units};
        
        \draw[draw, rounded corners=5pt, fill=fccolor] (1, -0.1) rectangle ++(2, 1) node[midway, draw=none, fill=none, align=center] (fc1_2) {FC layer\\10 units};
        
        \draw[draw, rounded corners=5pt, fill=conccolor] (3.5, 0.9) rectangle ++(1.5, 0.8) node[midway, draw=none, fill=none, align=center] (conc) {Concat};
        
        \draw[draw, rounded corners=5pt, fill=fccolor, line width=0.7mm] (5.5, 0.8) rectangle ++(2, 1) node[midway, draw=none, fill=none, align=center] (fc2) {FC layer\\20 units};
        
        \draw[draw, rounded corners=5pt, fill=fccolor, line width=0.7mm] (8, 0.8) rectangle ++(2, 1) node[midway, draw=none, fill=none, align=center] (fc3) {FC layer\\20 units};
        
        \draw[draw, rounded corners=5pt, fill=fccolor] (10.5, 0.8) rectangle ++(2, 1) node[midway, draw=none, fill=none, align=center] (fc4) {FC layer\\1 unit};
        
        \node[fill=outcolor] (output) at (13, 1.3) {};
        \node[draw=none, fill=none, rotate=0] at (13.1, 1.77) {output};

        \draw[->](inputmean1) -- (fc1_1);
        \draw[->](inputmean2) -- (fc1_1);
        \draw[->](inputstd1) -- (fc1_2);
        \draw[->](inputstd2) -- (fc1_2);
        \draw[->](fc1_1) -- (conc);
        \draw[->](fc1_2) -- (conc);
        \draw[->](conc) -- (fc2);
        \draw[->](fc2) -- (fc3);
        \draw[->](fc3) -- (fc4);
        \draw[->](fc4) -- (output);

    \end{tikzpicture}
    \caption{Diagram of the MLP for the Two Moons dataset.
        The mean and std input pass through two parallel fully-connected (``FC'') layers of 10 units, whose output is concatenated.
        Then, two 20-units fully-connected layers and the final classification layer are applied, which produce the final output.
        The two 20-units layers (depicted with bold borders) are made Bayesian---depending on the specific technique used.
    }
    \label{fig:mlp_twomoons}
\end{figure*}
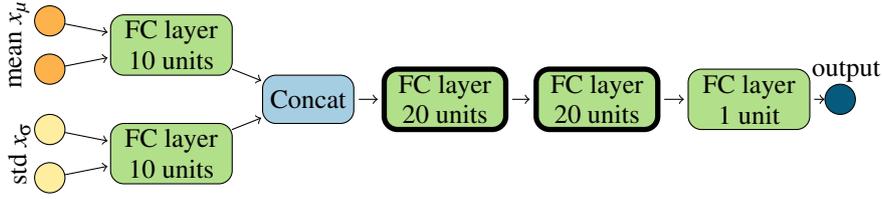

\textbf{Toy Regression}.
For a regression setting we use a commonly used sinusoid with variable amplitude and both homoscedatic ($\epsilon_2$) and heteroscedatic ($\epsilon_1$) aleatoric uncertainty, defined by:
\begin{equation}
    f(x) = x \sin(x) + \epsilon_1 x + \epsilon_2
\end{equation}
Where $\epsilon_1, \epsilon_2 \sim \mathcal{N}(0, 0.3)$. We produce 1000 samples for $x \in [0, 10]$ as a training set, and an out-of-distribution dataset is built with 200 samples for $x \in [10, 15]$.

\subsection{Predictive Uncertainty in NNs}\label{sec:predictive_uncertainty}

As previously stated, there is no direct way to compute predictive uncertainty on \emph{standard} deterministic NNs for regression.
In the case of $c$-way classification, instead, the output $f_\theta(x)\doteq \hat{y}$ is a vector of $c$ scalars, each scalar representing the \emph{confidence} that the model assigns to the input $x$ belonging to the corresponding category.
If softmax is applied to the output, we can see it as a probability distribution, and we can define two notions of uncertainty:
\begin{itemize}
    \item Entropy of the distribution (the \emph{flatter} the distribution, the more the model is uncertain)
    \begin{equation*}
        H(\hat{y}) = -\sum_{k=1}^c \hat{y}_k\log\hat{y}_k
    \end{equation*}
    \item Maximum of the distribution (the less confident the model is on assigning the model to the category with the maximum value, the more the model is uncertain).
    \begin{align}\label{eq:confidence}
        \text{Confidence}(\hat{y}) &= \max\{\hat{y}_k\} \\
        \text{Unconfidence} &= 1 - \text{Confidence}
    \end{align}
\end{itemize}
In the present work, we make use of both definitions of uncertainty. For a regression setting, we use the predictive mean $\mu(x)$ as a prediction and predictive standard deviation $\sigma(x)$ as uncertainty of that prediction.
\begin{equation}
    \mu(x) = M^{-1} \sum_i f_{\theta_i}(x)
\end{equation}
\begin{equation}
    \sigma^2(x) = M^{-1} \sum_i [f_{\theta_i}(x) - \mu(x)]^2
\end{equation}
Where $f_{\theta_i}$ is an stochastic bayesian model or ensemble members (via index $i$, see Section \ref{sec:bnns}) and $M$ is the number of forward passes or ensemble members, we usually use $M = 50$ for stochastic bayesian models.

In addition, the quality of the uncertainty estimates provided by the models can be assessed using \emph{calibration}.
The main idea is that, given a data point $x$, the model confidence should correspond to the accuracy attained on $x$.
By gathering the results on confidence and accuracy on a dataset, the confidence values can be divided in $B$ bins.
Then, the mean accuracy on each bin can be computed.
Given a bin $b$, we call $\text{confidence}_b$ the reference confidence on the bin; $\text{accuracy}_b$ is then the mean accuracy value.
Finally, the calibration can be measured by means of the ECE:
\begin{equation*}
    ECE = \sum_{b=1}^{B} N_b \frac{|\text{confidence}_b - \text{accuracy}_b|}{N}
\end{equation*}
where $N$ is the size of the dataset, and $N_b$ indicates the number of data points belonging to bin $b$.

\begin{figure*}
    \centering
    \scalebox{0.85}{
        \begin{tikzpicture}[
            node distance=0mm,
            box/.style args = {#1/#2}{shape=rectangle,
                text width=#1mm, minimum height=#2mm,
                draw, thick, inner sep=0pt, outer sep=0pt, 
                align=center, text=black}
            ]
            
            \node[draw=none, fill=none, rotate=90] at (-0.65, 0.75) {mean $x_\mu$};
            \node[draw=none, fill=none, rotate=90] at (-0.65, -0.75) {std $x_\sigma$};

            \node at (0,0.75) (input1) {\includegraphics[width=1cm]{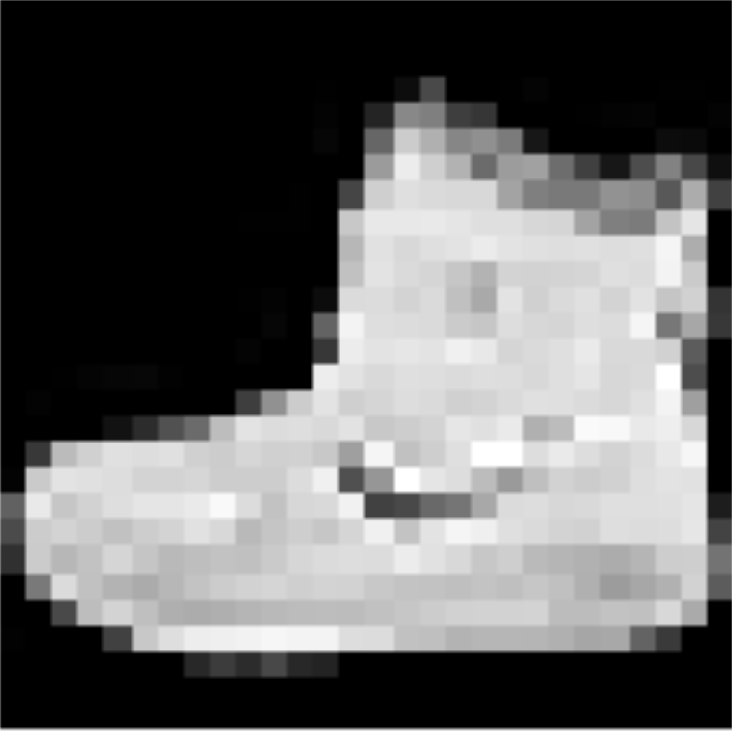}};
            
            \node at (0,-0.75) (input2) {\includegraphics[width=1cm]{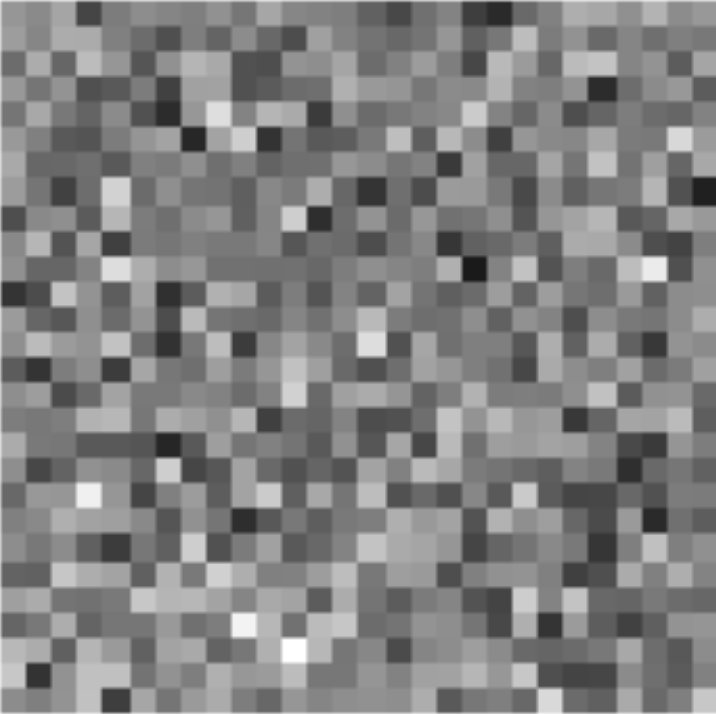}};
            
            \node[box=20/10, rounded corners=5pt, fill=yellow] (conv1_1) at (2, 0.75) {Conv $7\times 7$ \\ 32 ch., 2 str.};
            
            \node[box=20/10, rounded corners=5pt, fill=yellow] (conv1_2) at (2, -0.75) {Conv $7\times 7$ \\ 32 ch., 2 str.};
            
            \node[box=10/6, rounded corners=5pt, fill=conccolor] (concat) at (4, 0) {Concat};
            
            \node[box=20/10, rounded corners=5pt, rotate=90, fill=convcolor] (res1) at (5.5, 0) {\footnotesize Preact res.~block 64 ch.};
            
            \node[box=20/10, rounded corners=5pt, rotate=90, fill=convcolor] (res2) at (6.85, 0) {\footnotesize Preact res.~block 64 ch.};
            
            \node[box=20/10, rounded corners=5pt, rotate=90, fill=convcolor] (res3) at (8.1, 0) {\scriptsize Preact res.~block 128 ch., \scriptsize downsample};
            
            \node[box=20/10, rounded corners=5pt, rotate=90, fill=convcolor] (res4) at (9.45, 0) {\footnotesize Preact res.~block 128 ch.};
            
            \node[box=20/10, rounded corners=5pt, rotate=90, fill=convcolor] (res5) at (10.8, 0) {\scriptsize Preact res.~block 256 ch., \scriptsize downsample};
            
            \node[box=20/10, rounded corners=5pt, rotate=90, fill=convcolor] (res6) at (12.15, 0) {\footnotesize Preact res.~block 256 ch.};
            
            \node[box=20/10, rounded corners=5pt, rotate=90, fill=convcolor] (res7) at (13.5, 0) {\scriptsize Preact res.~block 512 ch., \scriptsize downsample};
            
            \node[box=20/10, rounded corners=5pt, rotate=90, fill=convcolor, line width=0.8mm] (res8) at (14.85, 0) {\footnotesize Preact res.~block 512 ch.};
            
            \node[box=20/6, rounded corners=5pt, rotate=90, fill=fccolor] (gap) at (16.0, 0) {GAP \& flatten};
            
            \node[circle,draw, minimum size=0.2cm, fill=meancolor] (n1) at (16.6,0.8) {};
            \node[circle,draw, minimum size=0.2cm, fill=meancolor] (n2) at (16.6,0.4) {};
            \node[rotate=90] (dots) at (16.6,0) {\scriptsize$\dots$};
            \node[] at (16.6, 1.3) {512};
            \node[circle,draw, minimum size=0.2cm, fill=meancolor] (n3) at (16.6,-0.4) {};
            \node[circle,draw, minimum size=0.2cm, fill=meancolor] (n4) at (16.6,-0.8) {};
            
            \foreach \x in {1, ..., 10}
            \node[circle, draw, minimum size=0.2cm, fill=outcolor] (output\x) at (17.75,{ (\x-1)*0.4-1.8}) {};
            \node[] at (17.75, 2.2) {10};

            \draw[->](input1) -- (conv1_1);
            \draw[->](input2) -- (conv1_2);
            \draw[->](conv1_1) -- (concat);
            \draw[->](conv1_2) -- (concat);
            \draw[->](concat) -- (res1);
            \draw[->](res1) -- (res2);
            \draw[->](res2) -- (res3);
            \draw[->](res3) -- (res4);
            \draw[->](res4) -- (res5);
            \draw[->](res5) -- (res6);
            \draw[->](res6) -- (res7);
            \draw[->](res7) -- (res8);
            \draw[->](res8) -- (gap);
            
            \foreach \x in {1,...,4}
            \foreach \y in {1, ..., 10}
            \draw[->, line width=0.01mm](n\x) -- (output\y);

        \end{tikzpicture}
    }
    \caption{
        Diagram depicting the two-input Preact-ResNet18 used on Fashion-MNIST.
        The input mean and standard deviation are passed through two $7\times 7$ convolutions with 32 channels and stride 2, whose outputs are concatenated.
        The data is then passed sequentially through a series of residual blocks (``Preact res. block'') with increasing number of channels.
        Some blocks operate downsampling of the spatial dimensions.
        A detailed depiction of the residual blocks is shown in \Cref{fig:resblock}.
        Following the last residual block, global average pooling (``GAP'') is applied to return a vector of size 512.
        This vector is passed through a fully-connected layer which produces the final output of 10 units.
        The last residual block (depicted with thick borders) can be rendered Bayesian by turning its convolutional layers into the corresponding Bayesian version, depending on the method used.
    }
    \vspace*{1em}
    \label{fig:preact18}
\end{figure*}
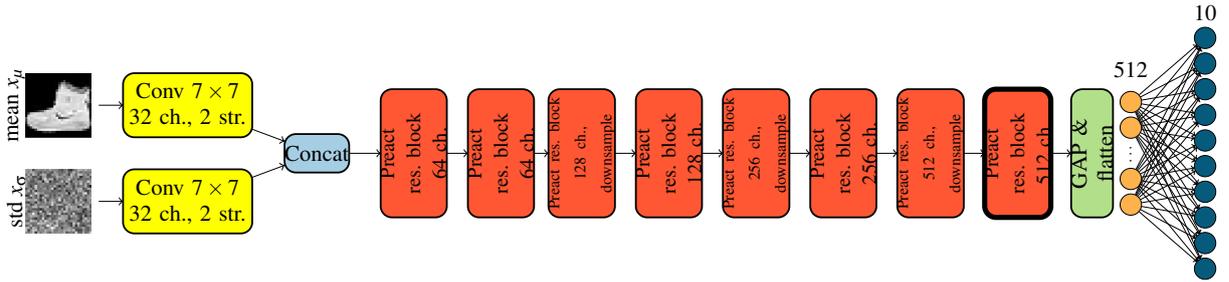

\subsection{Bayesian Neural Networks}\label{sec:bnns}

BNNs provide a paradigm shift, in which the parameters of the model are not scalars, but probability distributions.
This allows for a more reliable estimate of the predictive uncertainty \cite{naeini2015obtaining,ovadia2019trust} due to the more noisy nature of the prediction.
As in all Bayesian models, the driving principle behind BNNs is the computation of the posterior density $p(\theta|\mathcal{D})$, which is obtained via Bayes' theorem:
\begin{equation}\label{eq:bayes}
    p(\theta|\mathcal{D}) = \frac{\overbrace{p(\mathcal{D}|\theta)}^{\text{likelihood}}\cdot \overbrace{p(\theta)}^{\text{prior}}}{\underbrace{p(\mathcal{D})}_\text{marginal likelihood}}.
\end{equation}
The goal of Bayesian models is to start from a prior distribution defined on the parameters and updating the knowledge over these parameters by means of the evidence---the likelihood.
The updated probability distribution of the parameter is the posterior.
The computation of the marginal likelihood (the denominator in \Cref{eq:bayes}) is often computationally unfeasible, thus Bayesian Machine Learning often resort to approximations based on variations of Markov-Chain Monte-Carlo methods.
However, these methods are still too computationally demanding for BNNs \cite{blundell2015weight}, hence a number of techniques for approximating BNNs have been proposed in the last decade.
In the present work, we make use of a handful of these.

\textbf{MC-Dropout}.
MC-Dropout \cite{gal2016dropout} is a simple modification of the Dropout algorithm for NN regularization \cite{hinton2012improving}.
During the training phase, at each forward pass, some intermediate activations are randomly zeroed-out with a given probability value $p$.
During inference, the dropout behavior is turned off.
MC-Dropout maintains the dropout behavior active during the inference phase, thus allowing for the model to become stochastic.
A probability distribution over the output can hence be obtained by repeatedly running inference on the same data point---a process called \emph{sampling}.

\textbf{MC-DropConnect}.
DropConnect \cite{wan2013regularization} is a conceptual variation of Dropout: instead of suppressing activations, it acts by randomly zeroing-out some parameters with a given probability value $p$.
As for Dropout, DropConnect is also meant as a regularization technique to be activated during training.
MC-DropConnect, analogously to MC-Dropout, allows this method to be active also during inference, hence making the model stochastic.

\textbf{Direct Uncertainty Quantification}.
DUQ \cite{amersfoort2020uncertainty} is a method for creating a deterministic NN which incorporates reliable uncertainty estimates in its prediction.
It is designed only for classification tasks.
Its main idea is to redefine the final classification layer:
instead of a vector of $c$ scalars, the model thus produces $c$ embeddings in the same space $\mathbb{R}^m$.
The model is trained to pull the embeddings of the same categories closer to each other: the goal is to produce $c$ clusters corresponding to the classes.
The data point $x$ is then assigned to the category whose corresponding cluster centroid is nearest; similarly, uncertainty can be defined as the RBF distance to the nearest cluster centroid $\mu_k$:
\begin{equation}\label{eq:unc_duq}
    \text{Uncertainty}_{DUQ} = \max_{k\in\{1,\dots,c\}} \exp\left[  \frac{\frac{1}{m}|| f_\theta(x) - \mu_k ||^2_2}{2 \sigma^2}\right],
\end{equation}
with $\sigma$ being a hyperparameter.
\cite{amersfoort2020uncertainty} suggest to train DUQ models using gradient penalty \cite{drucker1992improving}, a regularization method which rescales the gradient by a hyperparameter $\lambda$.

\textbf{Flipout}.
Bayes By Backprop is a Variational Inference--inspired technique introduced by \cite{blundell2015weight}.
It allows to directly model the parameters of a BNN as Gaussian distributions, while introducing a technique to enable the backpropagation-based training typical of deterministic NNs.
It can be seen as a proper Bayesian method, since it directly models the probability distribution of the parameters, which are explicitly given a prior distribution.
The authors propose to use, as prior, a mixture of two zero-mean Gaussian with standard deviations $\sigma_1$ and $\sigma_2$ respectively and a mixture weight $\pi$.
BayesByBackprop makes use of a variational loss based on the Kullback-Leibler divergence between the approximate posterior learnt by the model and the true posterior.
BayesByBackprop is, though, computationally intensive and unstable; \cite{wen2018flipout} introduced a scheme, called \emph{Flipout}, to add perturbations to the training procedure, allowing to reduce training time and increasing stability.

\begin{table*}[t]
    \caption{
        Hyperparameters used in the implementation and training of the NNs.
        ``FMNIST'' is short for Fashion-MNIST and ``TM'' corresponds to the Two Moons dataset.
    }
    \centering
    \begin{tabular}{lllllllll} 
        & \multicolumn{2}{c}{\# epochs} & \multicolumn{2}{c}{Batch size} & \multicolumn{2}{c}{Other hyperparameters} & \multicolumn{2}{c}{\# samples for inference}  \\
        & TM & FMNIST & TM & FMNIST & TM & FMNIST & TM & FMNIST \\
        \toprule
        Deterministic NN & 100 & 15 & 32 & 256 &  &  & --- & --- \\
        \midrule
        MC-Dropout       & 100 & 15 & 32 & 256 & $p=0.2$ & $p=0.1$ & 100 & 25 \\
        MC-DropConnect   & 100 & 15 & 32 & 256 & \multicolumn{2}{l}{$p=0.05$} & 100 & 25 \\
        Ensembles         & 100 & 15 & 32 & 256 & \multicolumn{2}{l}{\#~components$=5$} & 5 & 5 \\
        DUQ              & 100 & --- & 32 & --- & $\sigma=0.1; \lambda=0.5$ & --- &  & --- \\
        Flipout          & 300 & 15 & 32 & 256 & \multicolumn{2}{l}{$\sigma_1=5; \sigma_2=2; \pi=0.5$} & 100 & 25 \\
        \bottomrule
    \end{tabular}
    \label{tab:hyperparams}
\end{table*}

\textbf{Ensembles}.
Ensembles are groups of non-Bayesian NNs with the same architecture and trained on the same data, but with different random initialization of their parameters.
They are not inherently Bayesian---their output is not stochastic---but the fact of having multiple outputs for a single data point allows us to make considerations on the predictive uncertainty.
Moreover, it has been shown \cite{lakshminarayanan2017simple} that ensembles are producing uncertainty estimates which are often superior in reliability to other methods here presented.

\subsection{Two-input NNs for Input Uncertainty}

In the \emph{deterministic} paradigm for NNs, the input is evaluated one-by-one, i.e., the data is passed one sample at the time without explicitly passing input uncertainty as input to the model.
Given the data space $\mathrm{R}^d$, the model is hence seen as a function $f_\theta:\mathrm{R}^d \rightarrow \mathcal{Y}\subseteq \mathcal{R}^k$, where $k$ is dependent on the task that the model needs to solve and $\theta$ indicates the parameters of the model (which are probability distributions in the case of BNNs).
In graphical terms, for both deterministic and Bayesian NNs, the model is represented with an \emph{input layer} with $n$ neurons.

In the present work, instead, inspired by \cite{rodrigues2023information}, we take a different approach and craft a NN architecture, which we call \emph{two-input NN}.
As the name suggests, this model has two input channels:
\begin{enumerate*}[label=(\alph*)]
    \item the \emph{mean} data $x_\mu$ (of dimension $d$), and
    \item the \emph{standard deviation} of the data $x_\sigma$, also of dimension $d$.
\end{enumerate*}
Thus, a two-input NN is represented as a function $f_\theta:\mathrm{R}^d\times\mathrm{R}^d \rightarrow \mathcal{Y}$.
This setting allows the NN to directly model input uncertainty.
We can see this process as feeding \emph{multiple versions} of the same data to the model, by accounting for the uncertainty---encoded in the standard deviation---which is intrinsic in the process of capturing this data.

We created three different versions of two-input NNs, one per dataset.

\textbf{NN for Two Moons}.
For the Two Moons dataset, we make use of a Multilayer Perceptron (MLP) with four input neurons (2 neurons for $x_\mu$, 2 neurons for $x_\sigma$) and three hidden layers of, respectively, \numlist{10;20;20} hidden units and ReLU activation and a final, one-unit classification layer.
The first hidden layer is duplicated so that the information of the mean and standard deviation flows parallely through them, after which the outputs are concatenated.
The two 20-units layers are all Bayesian, which means they implement either MC-Dropout, MC-DropConnect, or Flipout.
Ensembles and DUQ use regular MLPs, with DUQ replacing the classification layer with its custom implementation mentioned in \Cref{sec:bnns}.
A diagram of the architecture is depicted in \Cref{fig:mlp_twomoons}.

\textbf{NN for Fashion-MNIST}.
Inspired by \cite{harris2020fmix}, for Fashion-MNIST we use a custom Preact-ResNet18 \cite{he2016identity} with two modifications with respect to the original implementation.
\begin{enumerate}
    \item We turn this model into a two-input NN by modifying the first convolutional layer.
    Instead of a convolution with 64 output channels, we operate two convolutions in parallel with 32 output channels: the first one operates on $x_\mu$, the second one on $x_\sigma$.
    \item The second modification instead turns the NN into a BNN: we modify the two convolutional layers of the last residual block by implementing MC-Dropout, MC-DropConnect, or Flipout on them.
    Notice that the ensemble uses regular convolutions.
    Due to computational constraints, we don't train a model with DUQ for Fashion-MNIST.
\end{enumerate}

\textbf{NN for Toy Regression}.
We use a similar architecture than the NN for two moons. A MLP with four input neurons (2 neurons for $x_\mu$, 2 neurons for $x_\sigma$) and four hidden layers of, respectively, \numlist{10;10;20;20} hidden units and ReLU activation and a final, one-unit regression layer. There are separate hidden layers for input mean and input standard deviation, concatenated to connect to the final set of hidden layers.

\begin{figure*}[t]
    \centering
    \begin{subfigure}{0.49\textwidth}
        \centering
        \includegraphics[width=0.8\linewidth]{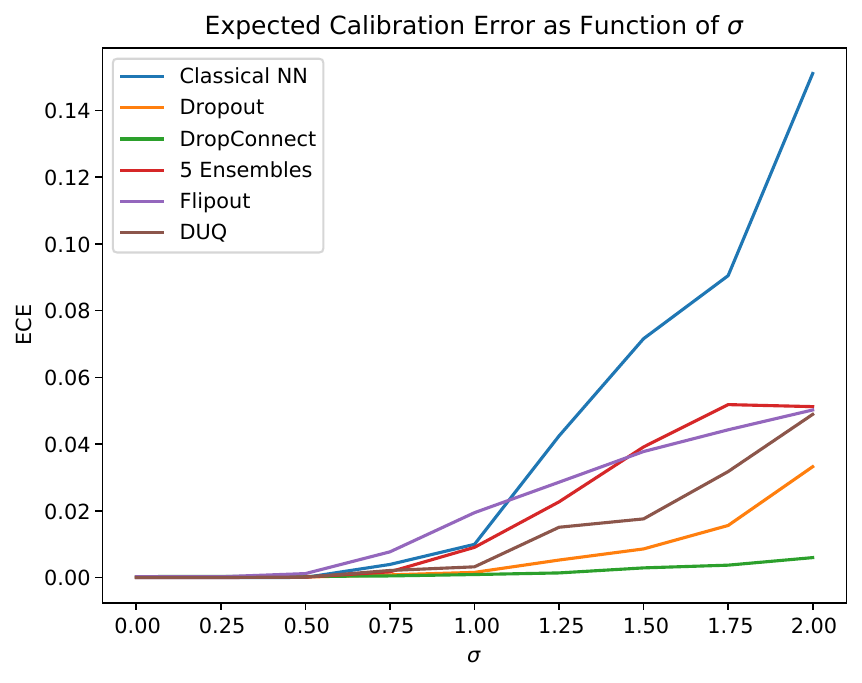}
        \caption{Expected Calibration Error}
    \end{subfigure}
    \begin{subfigure}{0.49\textwidth}
        \centering
        \includegraphics[width=0.8\linewidth]{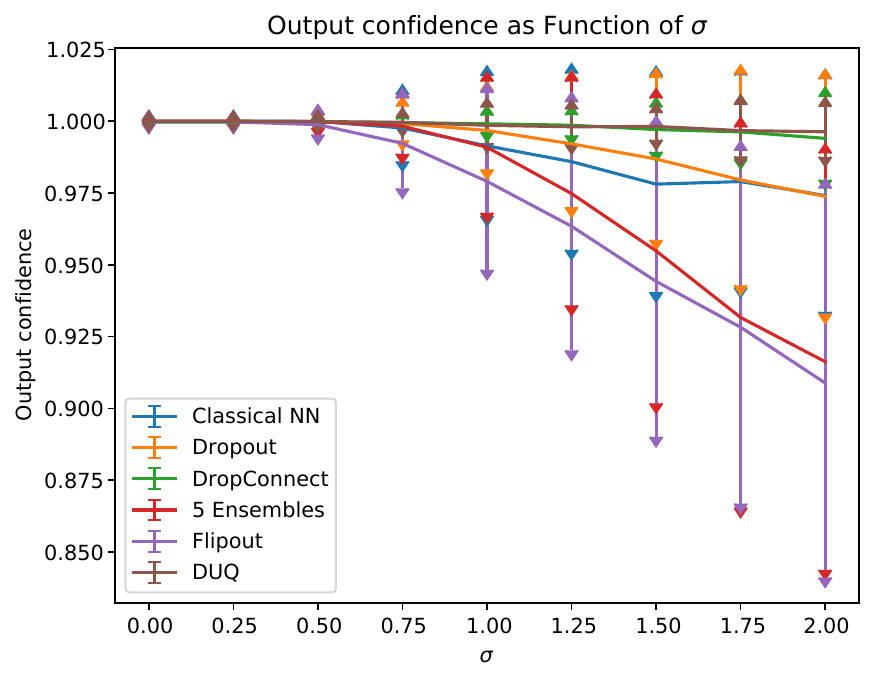}
        \caption{Output confidence as function of input uncertainty $\sigma$.}
    \end{subfigure}
    \vspace*{0.3em}
    \caption{Comparison of Expected Calibration Error and Output Confidences on the Two Moons dataset as input uncertainty $\sigma$ varies. The smallest variation in ECE is with DropConnect while Ensembles and Flipout have the largest decrease of output confidence.}
    \label{fig:two_moons_ece_inout}
\end{figure*}

\begin{figure*}[t]
    \begin{subfigure}{0.19\linewidth}
        \includegraphics[width=\textwidth]{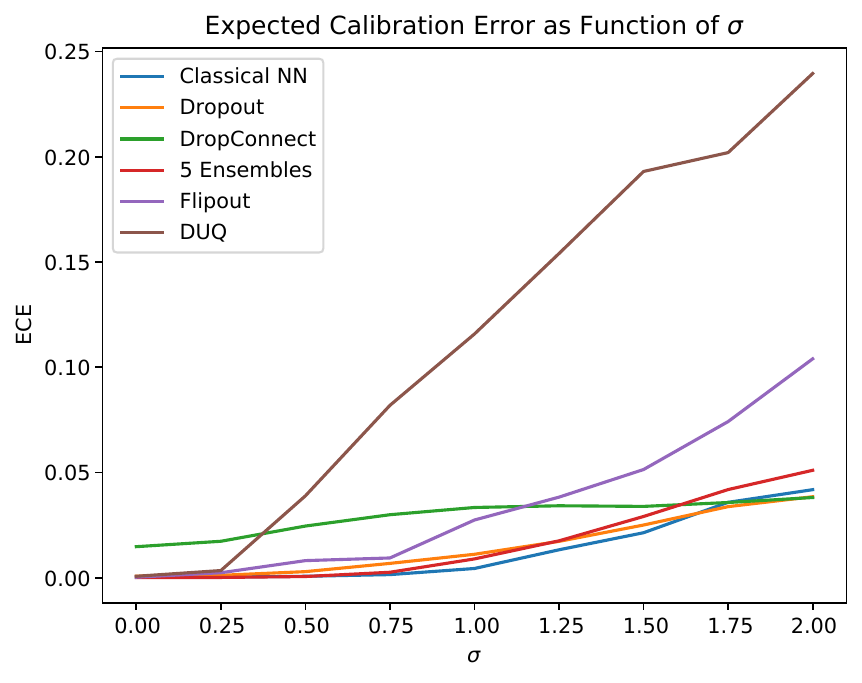}
    \end{subfigure}
    \begin{subfigure}{0.19\linewidth}
        \includegraphics[width=\textwidth]{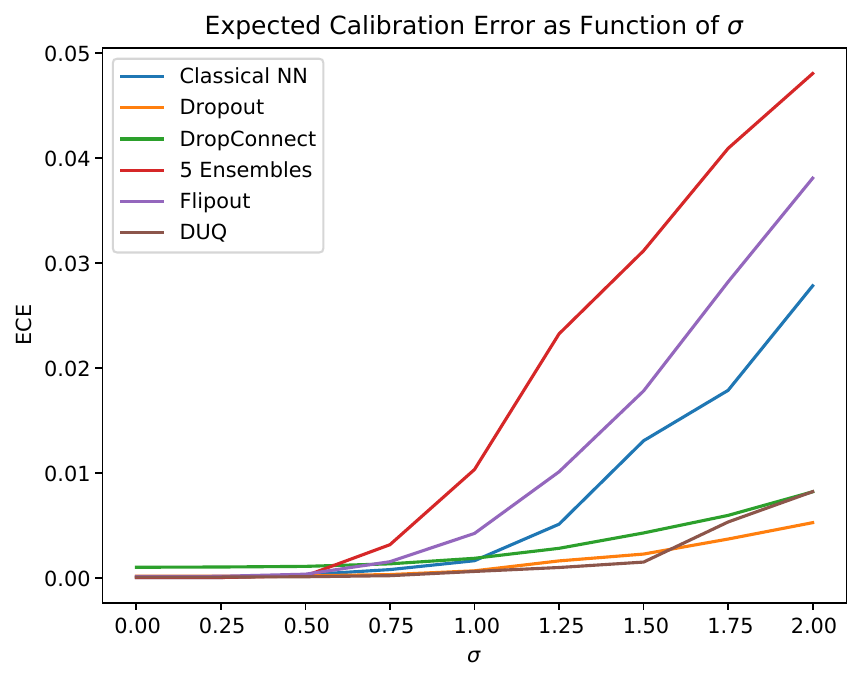}
    \end{subfigure}
    \begin{subfigure}{0.19\linewidth}
        \includegraphics[width=\textwidth]{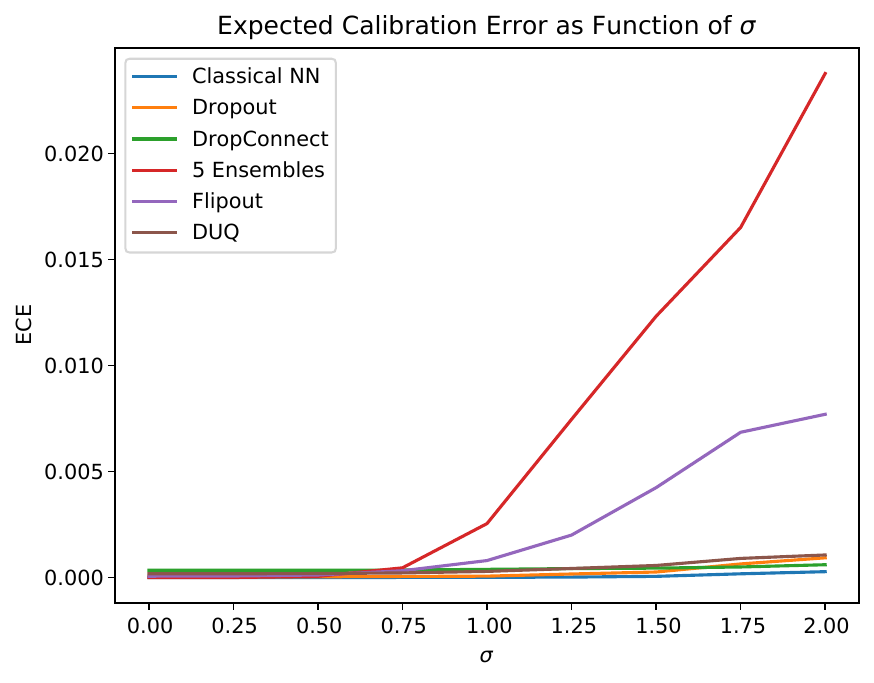}
    \end{subfigure}
    \begin{subfigure}{0.19\linewidth}
        \includegraphics[width=\textwidth]{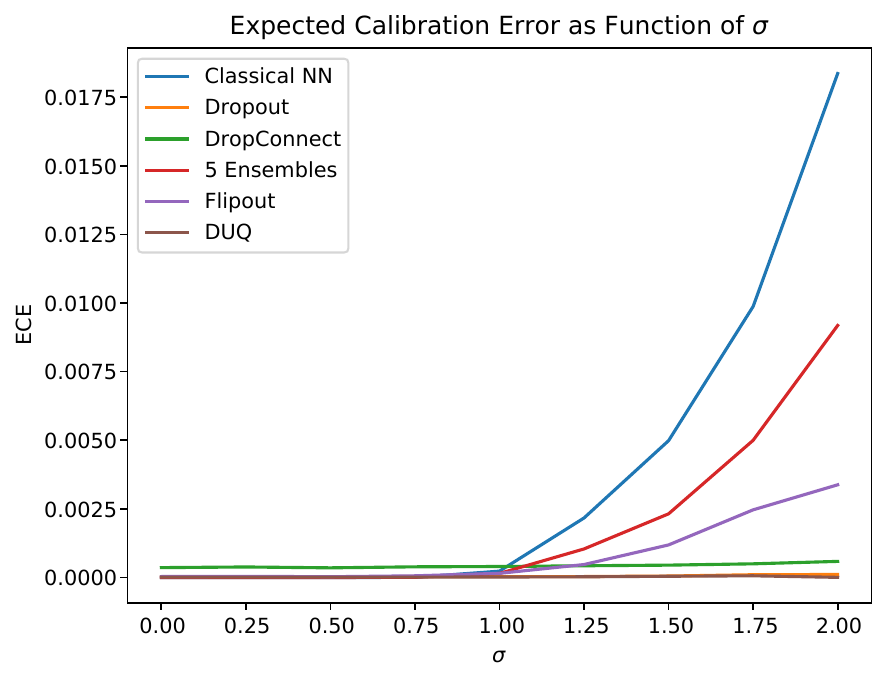}
    \end{subfigure}
    \begin{subfigure}{0.19\linewidth}
        \includegraphics[width=\textwidth]{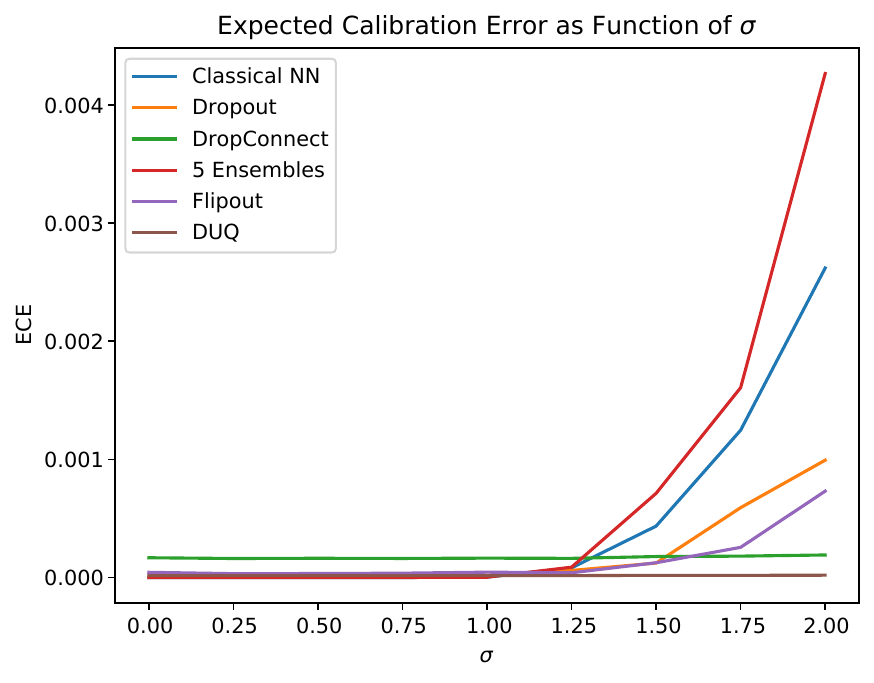}
    \end{subfigure}
    
    \begin{subfigure}{0.19\linewidth}
        \includegraphics[width=\textwidth]{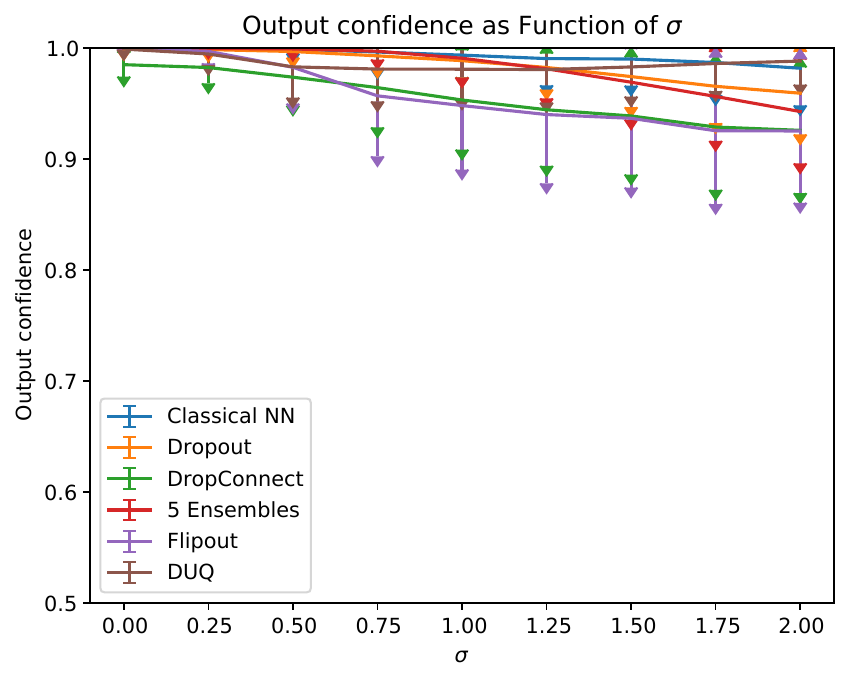}
        \caption{\tiny $\sigma=[0.0]$}
    \end{subfigure}
    \begin{subfigure}{0.19\linewidth}
        \includegraphics[width=\textwidth]{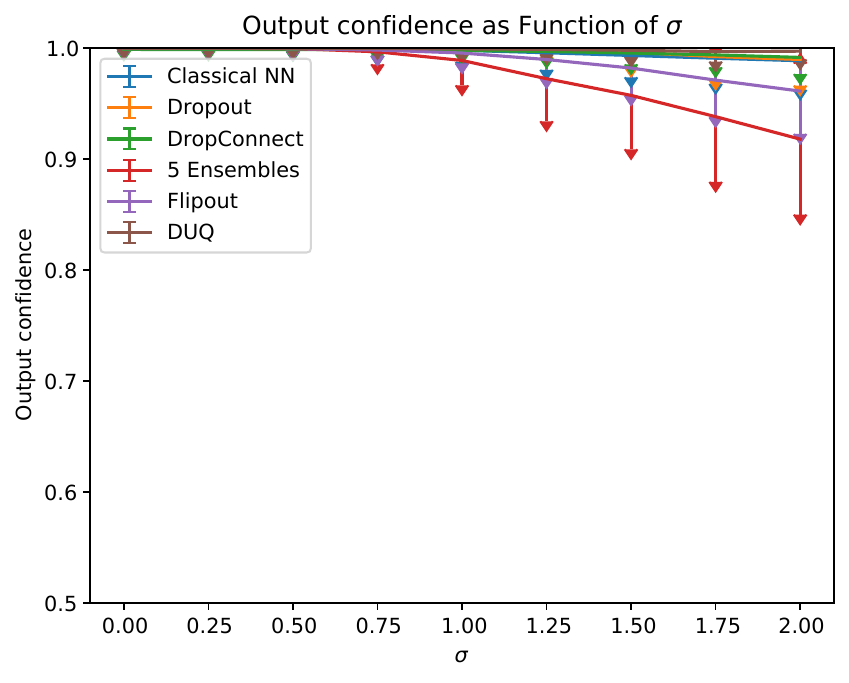}
        \caption{\tiny $\sigma=[0.0, 0.2]$}
    \end{subfigure}
    \begin{subfigure}{0.19\linewidth}
        \includegraphics[width=\textwidth]{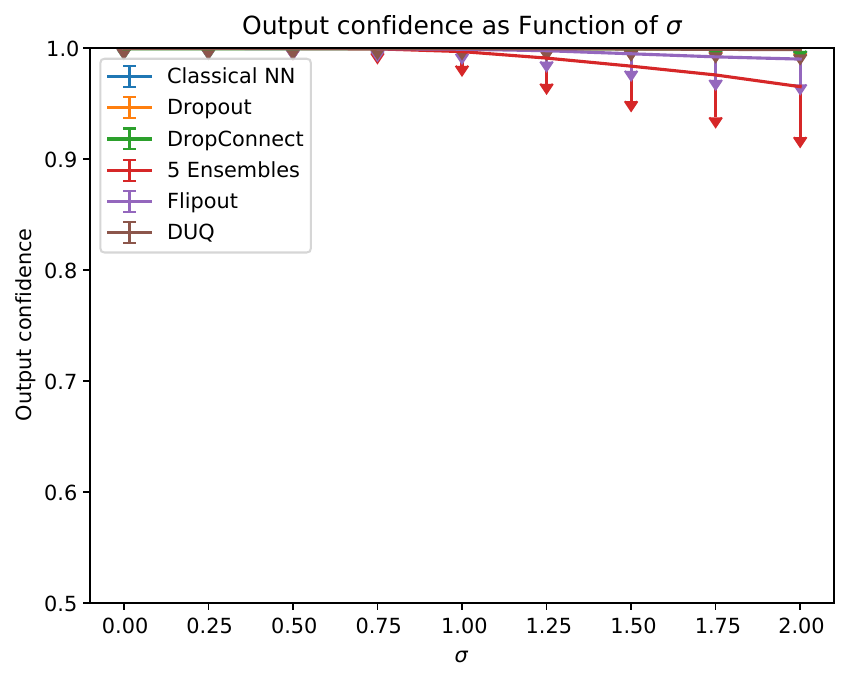}
        \caption{\tiny $\sigma=[0.0, 0.2, 0.4]$}
    \end{subfigure}
    \begin{subfigure}{0.19\linewidth}
        \includegraphics[width=\textwidth]{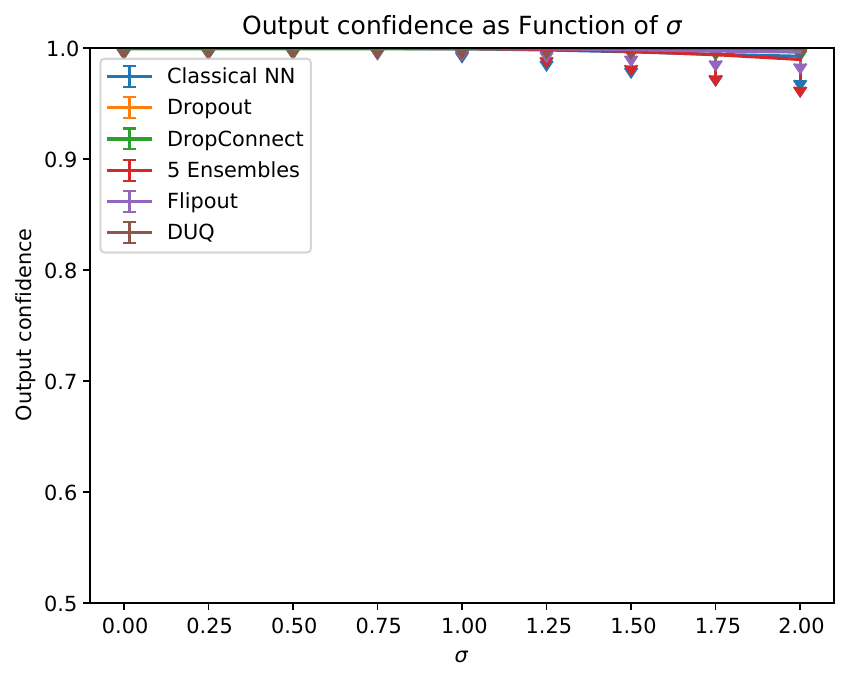}
        \caption{\tiny $\sigma=[0.0, 0.2, 0.4, 0.6]$}
    \end{subfigure}
    \begin{subfigure}{0.19\linewidth}
        \includegraphics[width=\textwidth]{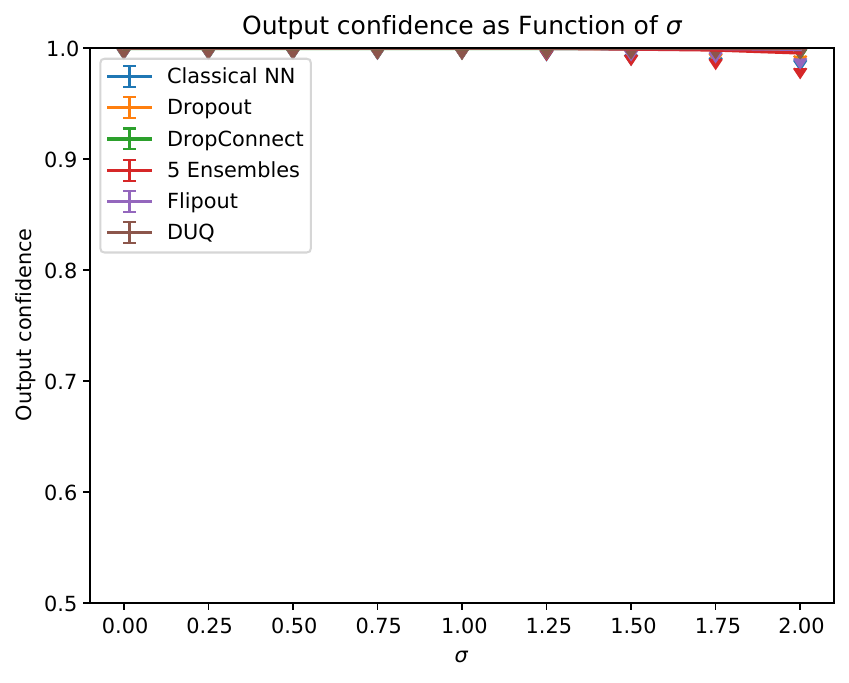}
        \caption{\tiny $\sigma=[0.0, 0.2, 0.4, 0.6, 0.8]$}
    \end{subfigure}
    \vspace*{2em}
    \caption{Results for the Two Moons dataset, setting when training set contains multiple values of sigma. ECE (top) and input/output uncertainty (bottom) are compared. Training on additional $\sigma$ values increases generalization for testing $\sigma > 1.0$, but makes most models except Ensembles to be insensitive to input uncertainty $\sigma$ by producing high confidences.}
    \label{fig:multisigma_two_moons_comparison}
\end{figure*}

\subsection{Experimental Setup}
We implement the models mentioned in the previous sections on Python, making use of the Keras library with Tensorflow backend.
For the Bayesian layers, we utilize Keras-Uncertainty \cite{valdenegro-toro2021keras}.
For the dataset Two moons, we run our experiments with all of the approximate BNN methods we introduced.
Due to computational reasons, we do not train a NN with DUQ on Fashion-MNIST.
In addition, for both datasets, we train a deterministic NN to allow for comparing results with respect to the frequentist setting.

The hyperparameters used for the implementation and training are showcased in \Cref{tab:hyperparams}.
In addition to what there indicated, we trained all of the models using the Adam optimizer \cite{kingma2014adam} with the Keras-default hyperparameters (learning rate of 0.001, $\beta_1$ 9f 0.9, and $\beta_2$ of 0.999).

For what concerns the injection of noise in the images, we simulate the process by passing the original data point in the $x_\mu$ input.
For the standard deviation input, we pass a structure $x_\sigma$ with the same size of the $x_\mu$ sampled from a normal distribution $x_\sigma \sim N(0, \sigma)$ with a given input noise standard deviation $\sigma$.
For Two moons, we fix $\sigma$ at 0.5.
For Fashion-MNIST, instead, we first normalize the images in the 0-1 range, then we generate the normal noise with $\sigma = 0.1$.

\begin{figure*}
    \centering
    \begin{subfigure}{0.49\textwidth}
        \centering
        \includegraphics[width=0.9\linewidth]{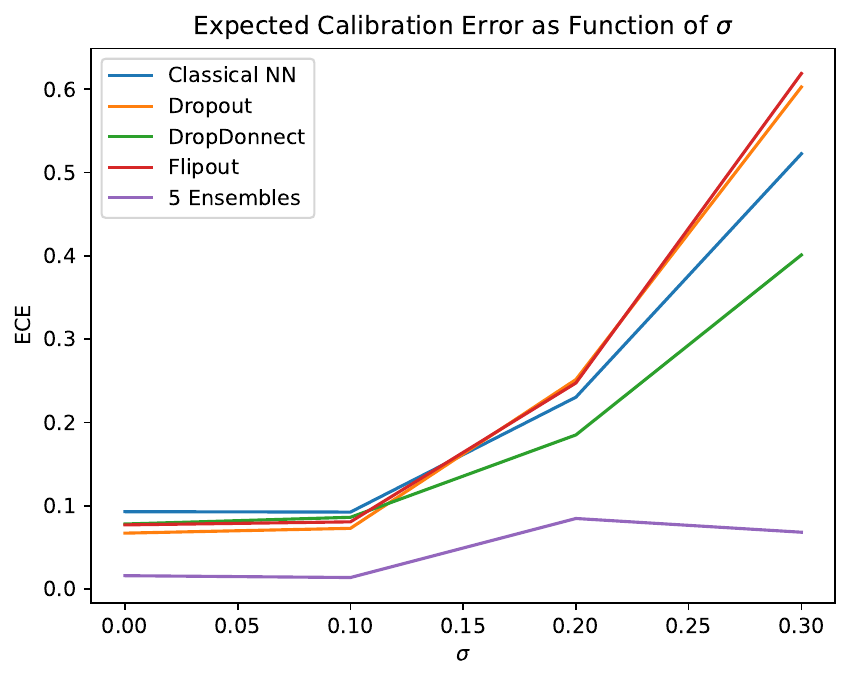}
        \caption{Expected Calibration Error}
    \end{subfigure}
    \begin{subfigure}{0.49\textwidth}
        \centering
        \includegraphics[width=0.9\linewidth]{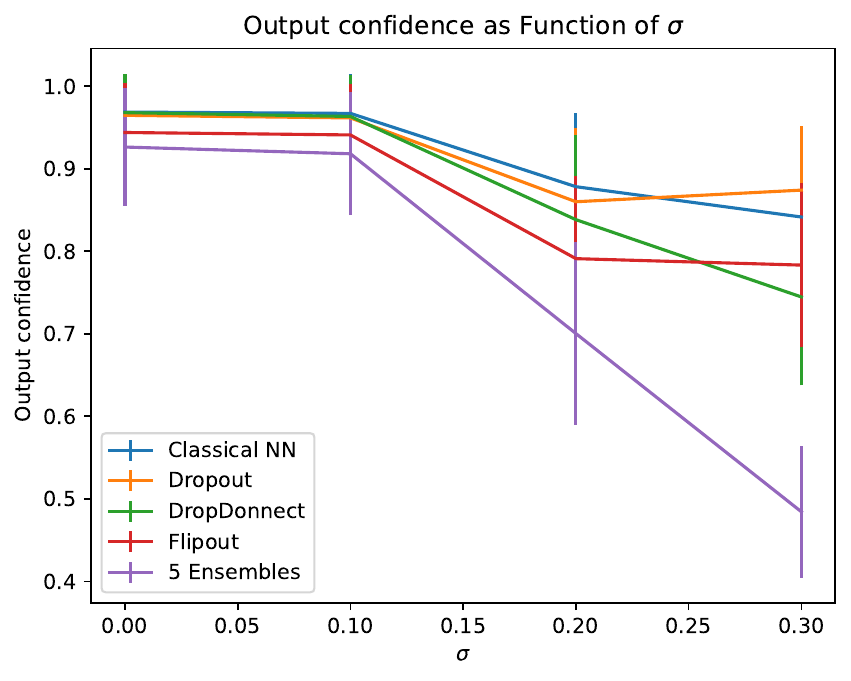}
        \caption{Output confidence as function of input uncertainty $\sigma$.}
    \end{subfigure}
    \vspace*{0.3em}
    \caption{Comparison of Expected Calibration Error and Output confidences for Fashion MNIST as input uncertainty $\sigma$ is varied. Note how Ensembles has little variation in calibration error and the largest decrease in confidence with increasing $\sigma$.}
    \label{fig:fmnist_ece_inout}
\end{figure*}

\textbf{Evaluation of Uncertainty}. For what concerns the evaluation of uncertainty, we test our models with increasing values of Gaussian noise.
We then provide two uncertainty-related comparison:
\begin{enumerate}[label=(\alph*)]
    \item ECE in function of $\sigma$. For DUQ, ECE is calculated considering the specific $\text{Uncertainty}_{DUQ}$ metric from \Cref{eq:unc_duq}.
    \item Output confidence in function of $\sigma$.
    The confidence is computed as per \Cref{eq:confidence}.
    For DUQ, the confidence is computed as the complementary of the normalized uncertainty obtained from \Cref{eq:unc_duq}.
\end{enumerate}
Finally, due to the ease of visualization provided by the 2D nature of Two Moons, we plot the dataset and the uncertainty, calculated in terms of entropy, for a lattice of points around the dataset.

\section{Experimental Results}

We perform experiments on two datasets. The purpose of these experiments is to evaluate if a Bayesian neural network and other models with uncertainty estimation, can learn to model input uncertainty from two inputs (mean and standard deviation). We test this with a simple setup, we train models with fixed levels of input uncertainty, and then test with increasing levels of input uncertainty.

Our expectation is, if a model properly learns the relationship between input and output uncertainty, then increasing input uncertainty should lead to increases in output uncertainty. We measure output uncertainty via entropy and maximum softmax confidence, and quality of uncertainty via the expected calibration error.

\subsection{Two Moons Toy Example}

We first evaluate on a toy example, the Two Moons dataset, available in scikit-learn, as it allows for easy control of input uncertainty and to visualize its effects. We perform two experiments, first we train a model with a single $\sigma$ value during training, and then train a model with multiple $\sigma$ values.

We first examine the case for a single training uncertainty, we use $\sigma = 0.2$. We plot and compare the output entropy distribution over the input domain, keeping the mean fixed but varying the input uncertainty $\sigma$ from $\sigma = 0.0$ to $\sigma = 2.0$. These results are presented in Figure \ref{fig:two_moons_comparison} and detailed plots for two metrics in Figure \ref{fig:two_moons_ece_inout}.

These results show that only Ensembles and Flipout significantly decrease their output confidence as the input uncertainty $\sigma$ increases, while a classical NN without uncertainty estimation becomes highly miscalibrated, and other methods only produce minor decreases in output confidence. No variations in ECE and output confidence while $\sigma$ increases indicates that the model might be ignoring the input uncertainty, which is exactly the behavior we wanted to test.

We secondly examine the case for multiple training input uncertainties, using $\sigma \in [0.0, 0.2, 0.4, 0.6, 0.8]$, and testing with $\sigma \in [0.0, 0.25, 0.5, 0.75, 1.0, 1.25, 1.50, 1.75, 2.0]$ progressively. These results are presented in \Cref{fig:multisigma_two_moons_comparison}.

These results indicate that Flipout is always miscalibrated relative to other methods, and that all uncertainty estimation methods minus Flipout seem to be insensitive to input uncertainty, always producing high output uncertainty. At the end of the spectrum, training with five different $\sigma$ values (\Cref{fig:multisigma_two_moons_comparison}e), most methods have learned to ignore the input uncertainty as output confidence barely varies.

\subsection{Fashion-MNIST Image Classification}

We then proceed to evaluate our hypothesis on Fashion-MNIST.
We train the models on a fixed standard deviation value $\sigma=0.1$ and report the corresponding test-set accuracy in \Cref{tab:fmnist_accuracy}. ECE and output confidence (computed on the test-set) as function of input uncertainty are presented in \Cref{fig:fmnist_ece_inout}.
In this case, we restrict the range of standard deviation for the testing to $\sigma=$\numlist{0.0;0.1;0.2;0.3}
While we train on a single input $\sigma$, Ensembles and Flipout decrease their output confidence while input uncertainty $\sigma$ increases, as expected, while other methods do not. The results are similar to what we observed on the Two Moons dataset, indicating that our results and experiments generalize to a more complex image classification setting.

\begin{table}[t]
    \centering
    \caption{Train-- and test--set accuracy attained by our NNs trained on Fashion-MNIST.}
    \begin{tabular}{lll}
        \toprule
        Model & Train accuracy & Test accuracy \\
        \midrule
        Deterministic NN & 98.6\% & 88.6\%\\
        MC-Dropout       & 98.7\% & 88.7\%\\
        MC-DropConnect   & 98.7\% & 87.7\%\\
        Ensemble         & 98.5\% & 88.3\%\\
        Flipout          & 95.5\% & 85.9\%\\
        \bottomrule
    \end{tabular}
    
    \label{tab:fmnist_accuracy}
\end{table}

\begin{figure*}[t]
    \begin{center}
        \includegraphics[width=0.7\linewidth]{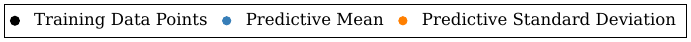}
    \end{center}
    \includegraphics[width=\linewidth]{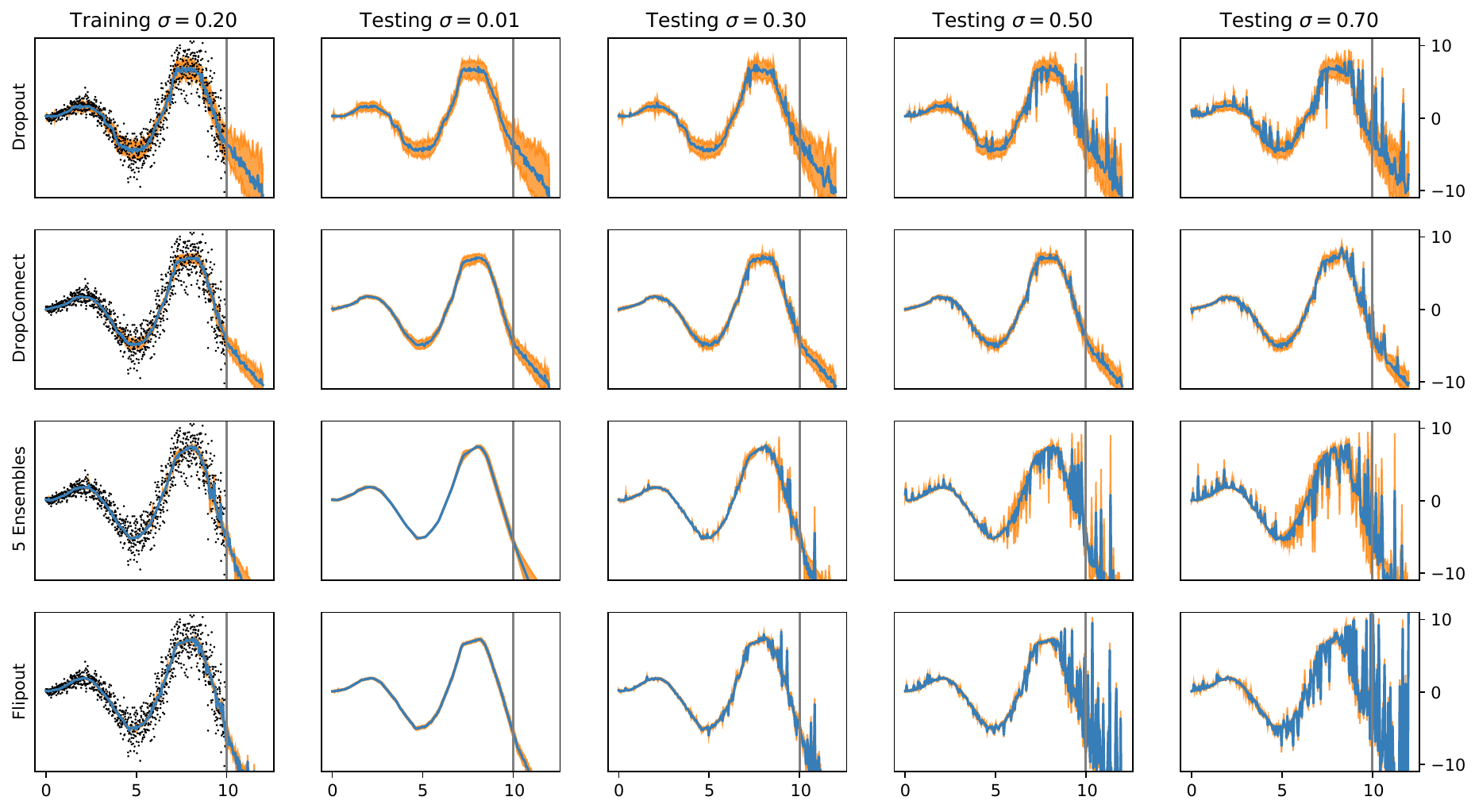}
    \caption{Comparison on a toy regression setting with training $\sigma = 0.2$ and variable testing standard deviation. Consistent with classification results, Ensembles and Dropout have the highest sensitivity to input uncertainty $\sigma$.}
    \label{fig:regression_comparison}
\end{figure*}

\begin{figure*}        
    \begin{subfigure}{0.24\linewidth}
        \includegraphics[width=\linewidth]{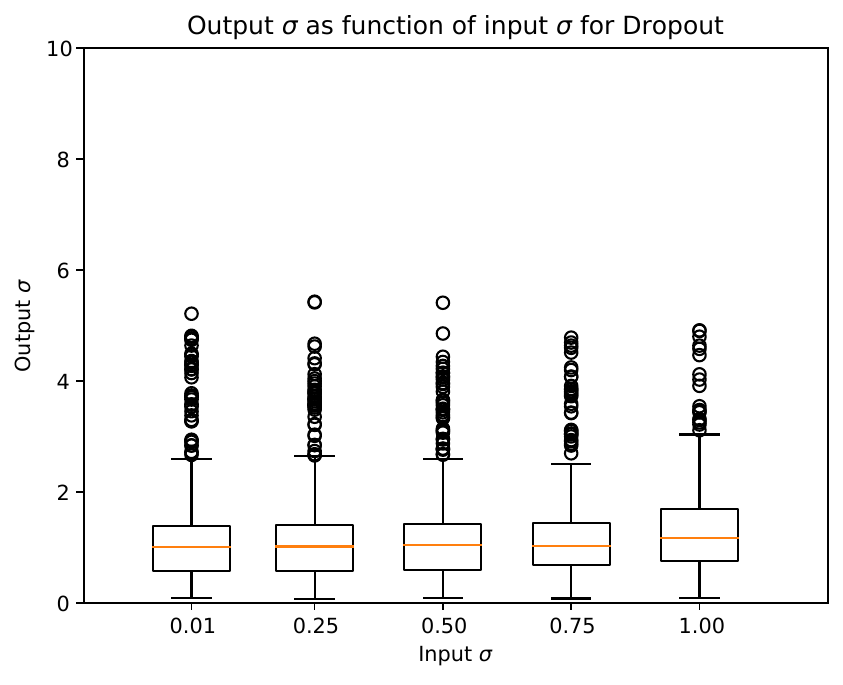}
        \caption{Dropout}
    \end{subfigure}
    \begin{subfigure}{0.24\linewidth}
        \includegraphics[width=\linewidth]{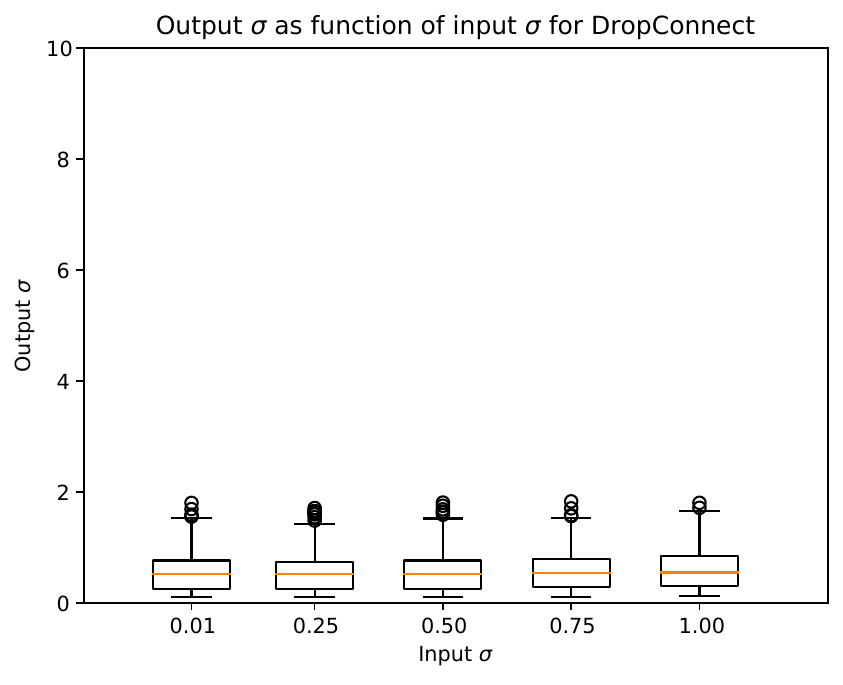}
        \caption{Dropconnect}
    \end{subfigure}
    \begin{subfigure}{0.24\linewidth}
        \includegraphics[width=\linewidth]{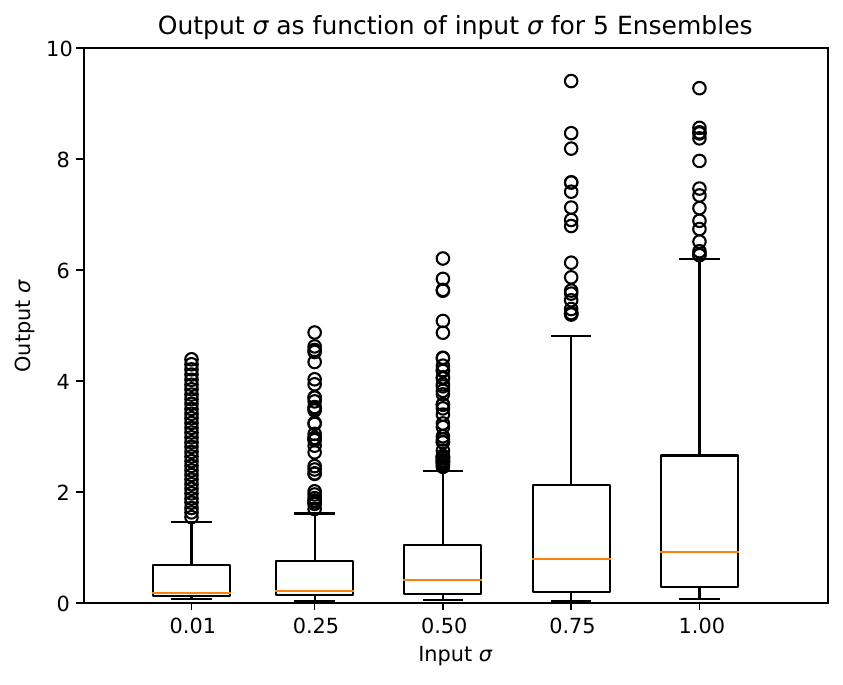}
        \caption{5 Ensembles}
    \end{subfigure}
    \begin{subfigure}{0.24\linewidth}
        \includegraphics[width=\linewidth]{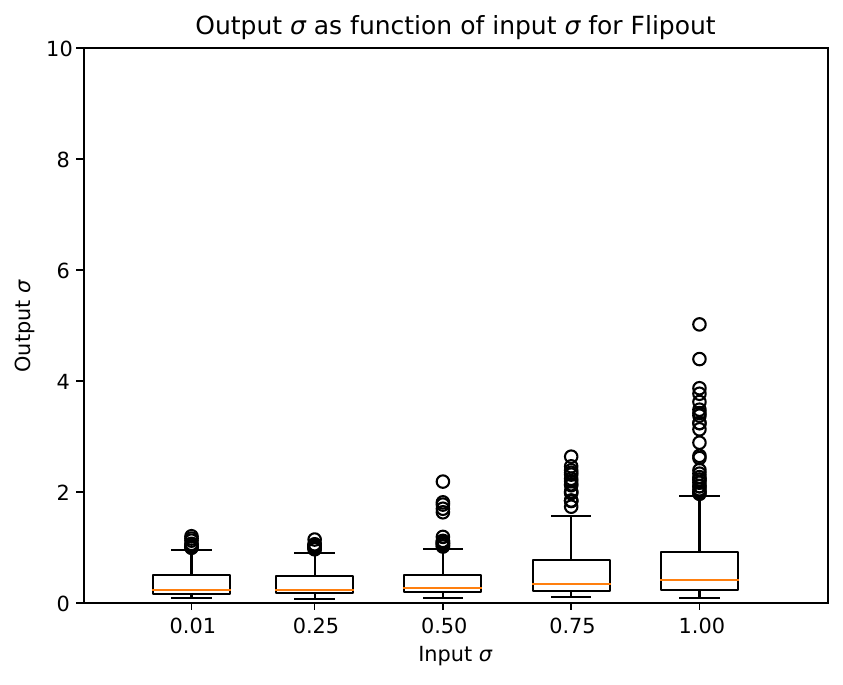}
        \caption{Flipout}
    \end{subfigure}
    \vspace*{0.3em}
    \caption{Comparison of output standard deviation as function of input standard deviation for the toy regression setting via boxplots. Flipout seems barely sensitive to changes in input uncertainty, while Ensembles has the highest reaction to increased input uncertainty.}
    \label{fig:regression_inout}
\end{figure*}

\subsection{Toy Regression Example}

Finally we evaluate results on the toy regression example, these results are shown in Figure \ref{fig:regression_comparison} in terms of predictions with epistemic uncertainty, and Figure \ref{fig:regression_inout} by comparing input and output uncertainties.

Dropout and DropConnect are insensitive to changes in input uncertainty, mostly by producing large uncertainties that do not vary with the input uncertainty, while Ensembles and Flipout do have varying output uncertainty with the input uncertainty, in a monotonic way, Flipout has increasing uncertainty mostly as variations in the predictive mean, while Ensembles has variation mostly on the standard deviation, so we consider the Ensemble results to be more representative of our expectations on how output uncertainty should behave as functions of input uncertainty.

Results on this regression example are consistent with our previous classification results, indicating that the results are general enough across tasks.

\section{\uppercase{Conclusions}}
\label{sec:conclusion}

In the present work, we investigated the quality of modeling aleatoric uncertainty by classical Neural Networks (NNs) and Bayesian Neural Networks (BNNs) when the input uncertainty is fed directly to the models in addition to the \emph{canonical} input.
We proposed a simple setting in which we artificially injected Gaussian noise in two famous benchmark datasets---Two Moons and Fashion-MNIST---often used in uncertainty estimation studies.
This simulates a natural environment in which the data are collected by means of sensors, which always exhibit a certain degree of noise.
For having the models receive the input uncertainty directly, while this being separated from the data itself, we crafted a set of NN architectures---which we dubbed \emph{two-input} NNs---with two input channels, one for the mean data and the other for the standard deviation corresponding to the added noise.
We trained these models on the above mentioned datasets, with a fixed level of noise, using five approximate BNN techniques: MC-Dropout, MC-DropConnect, Flipout, Ensembles, and Direct Uncertainty Quantification (DUQ).

We tested these models with data with none to high-levels of noise and proceeded to compute the output confidence and the Expected Calibration Error (ECE) as a function of noise. 
Our hypothesis was that, generically, these models would exhibit a certain degree of \emph{insensitivity} to added noise, where their confidence would still be high even when data with high noise---which are effectively out-of-distribution in the settings---are presented to them.
The results are pointing in this direction: both on Two Moons and on Fashion-MNIST, the output confidence for most of the methods remains high, while the Ensembles show a pronounced drop in confidence as the input uncertainty increases.
On the other hand, the results elicited by ECE are not conclusive, depicting a noisier scenario for what concerns the (mis)calibration of the models.

On Two Moons, where we conducted more extensive analyses, we noticed that, after injecting higher levels of noise in the training process, the models would essentially start ignoring the signal coming from the input uncertainty and always produce very confident predictions and being less miscalibrated.
Despite this seemingly being an optimal behavior, in which robustness to noise is enforced, can cause the NNs to fail at recognizing anomalous data, which is one of the reasons for adopting BNNs: by providing more reliable confidence estimates, confidence can be thresholded to filter out outliers and avoid classifying them.

Thus, our analyses suggest that both \emph{deterministic} NNs and BNNs fail, in a certain degree, to model data uncertainty when this one is provided explicitly as input, with ensembles---which are already known in the literature to being particularly powerful then other methods at producing good uncertainty estimates---and, to a lower extent, Flipout, showing the biggest drop in confidence when presented with very noisy inputs.

Our work, despite being the first analysis on the uncertainty of the NNs when directly modeling input uncertainty, is still quite small scale and mostly observational, and could potentially benefit for more extensive analyses.
For instance, larger-scale datasets might be used---although BNNs are notoriously difficult and slow to train on bigger datasets.
Also, we could extend the selection of BNN-training schemes to other methods, like the more recent SWAG \cite{maddox2019simple}, or Hamiltonian Monte--Carlo \cite{neal2011mcmc}, which is still considered the golden standard for Bayesian modeling, albeit very unfeasible to apply in the large-scale datasets used in modern Deep Learning.
Finally, our study could benefit from the addition on the analysis of \emph{uncertainty disentanglement} by the BNNs, to understand to what extent the models are able to integrate aleatoric uncertainty into the input uncertainty.

\bibliographystyle{apalike}
{\small
\bibliography{references}}

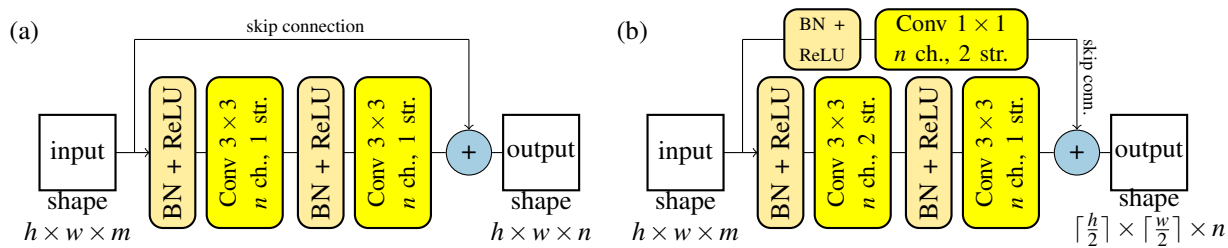
\begin{figure*}[bp]
    \centering
    \begin{tikzpicture}[
        node distance=0mm,
        box/.style args = {#1/#2}{shape=rectangle,
            text width=#1mm, minimum height=#2mm,
            draw, thick, inner sep=0pt, outer sep=0pt, 
            align=center, text=black}
        ]
        \node[] at (-0.7, 1.6) {(a)};
        \node[box=10/10, rotate=0, fill=white] (input) at (0, 0) {input};
        \node[align=center] at (0, -0.85) {shape\\$h\times w\times m$};
        
        \node[box=20/6, rounded corners=5pt, rotate=90, fill=stdcolor](bn1) at (1.25, 0) {BN + ReLU};
        \node[box=20/10, rounded corners=5pt, rotate=90, fill=yellow] (conv1) at (2.2, 0) {\footnotesize Conv $3\times 3$ \\ $n$ ch., 1 str.};
        
        \node[box=20/6, rounded corners=5pt, rotate=90, fill=stdcolor](bn2) at (3.2, 0) {BN + ReLU};
        \node[box=20/10, rounded corners=5pt, rotate=90, fill=yellow] (conv2) at (4.15, 0) {\footnotesize Conv $3\times 3$ \\ $n$ ch., 1 str.};
        \node[circle, draw, minimum size=0.2cm, fill=conccolor] (sum) at (5.15, 0) {+};
        \node[box=10/10, rotate=0, fill=white] (output) at (6.1, 0) {output};
        \node[align=center] at (6.1, -0.85) {shape\\$h\times w\times n$};
        
        \node[] at (3,1.65){\scriptsize skip connection};
        
        \coordinate (a) at ($(input)!0.6!(bn1)$);
        
        \draw[-](input) -- (a);
        \draw[->](a) -- (bn1);
        \draw[-](bn1) -- (conv1);
        \draw[-](conv1) -- (bn2);
        \draw[-](bn2) -- (conv2);
        \draw[-](conv2) -- (sum);
        \draw[-](sum) -- (output);
        
        \draw[-](a) -- ($(a) + (0,1.5)$);
        \draw[-]($(a) + (0,1.5)$) -- ($(sum) + (0,1.5)$);
        \draw[->]($(sum) + (0,1.5)$) -- (sum);

        \node[] at (7.3, 1.6) {(b)};
        \node[box=10/10, rotate=0, fill=white] (input2) at (8, 0) {input};
        \node[align=center] at (8, -0.85) {shape\\$h\times w\times m$};
        
        \node[box=20/6, rounded corners=5pt, rotate=90, fill=stdcolor](bn12) at (9.25, 0) {BN + ReLU};
        \node[box=20/10, rounded corners=5pt, rotate=90, fill=yellow] (conv12) at (10.2, 0) {\footnotesize Conv $3\times 3$ \\ $n$ ch., 2 str.};

        \node[box=20/6, rounded corners=5pt, rotate=90, fill=stdcolor](bn22) at (11.2, 0) {BN + ReLU};
        \node[box=20/10, rounded corners=5pt, rotate=90, fill=yellow] (conv22) at (12.15, 0) {\footnotesize Conv $3\times 3$ \\ $n$ ch., 1 str.};
        \node[circle, draw, minimum size=0.2cm, fill=conccolor] (sum2) at (13.15, 0) {+};
        \node[box=10/10, rotate=0, fill=white] (output2) at (14.1, 0) {output};
        \node[align=center] at (14.1, -0.85) {shape\\$\lceil\frac{h}{2}\rceil\times \lceil\frac{w}{2}\rceil \times n$};
        
        \node[box=10/8, rounded corners=5pt, fill=stdcolor] (bnres) at (9.8,1.5) {\scriptsize BN + ReLU};
        
        \node[box=20/8, rounded corners=5pt, fill=yellow] (convres) at (11.5,1.5) {\footnotesize Conv $1\times 1$ \\ $n$ ch., 2 str.};
        
        \node[rotate=270] at (13.3, 1.0){\scriptsize skip conn.};
        
        \coordinate (a2) at ($(input2)!0.6!(bn12)$);
        
        \draw[-](input2) -- (a2);
        \draw[->](a2) -- (bn12);
        \draw[-](bn12) -- (conv12);
        \draw[-](conv12) -- (bn22);
        \draw[-](bn22) -- (conv22);
        \draw[-](conv22) -- (sum2);
        \draw[-](sum2) -- (output2);
        
        \draw[-](a2) -- ($(a2) + (0,1.5)$);
        \draw[-]($(a2) + (0,1.5)$) -- (bnres);
        \draw[-](bnres) -- (convres);
        \draw[-](convres) -- ($(sum2) + (0,1.5)$);
        \draw[->]($(sum2) + (0,1.5)$) -- (sum2);
    \end{tikzpicture}
    \vspace*{0.5em}
    \caption{
        Diagrams depicting the two types of residual block used in the Preact-ResNet18 architecture:
        (a) standard residual block: a classic residual block with two $3\times 3$ convolutions (``Conv'') with a predefined number of output channels $n$, stride and padding of 1.
        The residual blocks are preceded by batch normalization (``BN'') and ReLU activation.
        The input and output have the same spatial dimensions $h$ and $w$.
        (b) residual block with downsample: it operates a downsampling on the spatial dimension by modifying the first convolution to have a stride of 2 instead of 1.
        In order to match the spatial dimension after the two convolutions, the skip connection presents a BN followed by ReLU and a $1\times 1$ convolution with stride 2 and padding 1.
        The output has spatial dimensions which are half the size of the input's.
    }
    \label{fig:resblock}
\end{figure*}

\end{document}